\documentclass{article}

\usepackage[preprint]{neurips_2026}

\usepackage[utf8]{inputenc}
\usepackage[T1]{fontenc}
\usepackage{microtype}
\usepackage{graphicx}
\usepackage{booktabs}
\usepackage{array}
\usepackage{algorithm}
\usepackage{algpseudocode}
\usepackage{amsmath}
\usepackage{amssymb}
\usepackage{amsfonts}
\usepackage{amsthm}
\usepackage{mathtools}
\usepackage{bm}
\usepackage{nicefrac}
\usepackage{xcolor}
\usepackage{wrapfig}
\usepackage{hyperref}
\usepackage[capitalize,nameinlink]{cleveref}
\hypersetup{hidelinks}
\hypersetup{pdftitle={Visual-Redundancy-Controlled Parallel Decoding for Diffusion-Based Multimodal Large Language Models},pdfauthor={Yulin Yuan, Hongshuo Zhao, Xiangming Meng}}

\crefname{section}{Sec.}{Secs.}
\Crefname{section}{Section}{Sections}
\crefname{table}{Tab.}{Tabs.}
\Crefname{table}{Table}{Tables}
\crefname{figure}{Fig.}{Figs.}
\Crefname{figure}{Figure}{Figures}
\crefname{equation}{Eq.}{Eqs.}
\Crefname{equation}{Equation}{Equations}

\newcommand{\mb}[1]{\mathbf{#1}}

\title{Visual-Redundancy-Controlled Parallel Decoding for Diffusion-Based Multimodal Large Language Models}

\author{%
  \normalfont\small
  \begin{tabular*}{0.97\textwidth}{@{\extracolsep{\fill}}ccc@{}}
    \parbox[t]{0.29\textwidth}{\centering
      \textbf{Yulin Yuan} \\
      Zhejiang University \\
      yulinyuan@zju.edu.cn
    }
    &
    \parbox[t]{0.31\textwidth}{\centering
      \textbf{Hongshuo Zhao} \\
      ZJU-UIUC Institute, \\
      Zhejiang University \\
      hongshuo.24@intl.zju.edu.cn
    }
    &
    \parbox[t]{0.31\textwidth}{\centering
      \textbf{Xiangming Meng\thanks{Corresponding author.}} \\
      ZJU-UIUC Institute, \\
      Zhejiang University \\
      xiangmingmeng@intl.zju.edu.cn
    }
  \end{tabular*}
}

\begin{document}

\maketitle

\begin{abstract}
Diffusion-based multimodal large language models (dMLLMs) decode by iteratively predicting tokens at multiple masked positions in parallel. This turns each decoding step into a position-selection problem: the model must choose not only which predictions are reliable in isolation, but also which positions should be committed together as context for later decoding steps. Confidence-based unmasking ranks masked positions independently and commits the top-$K$ positions, largely ignoring whether the committed tokens provide complementary visual grounding. We identify a step-level limitation of this strategy in multimodal settings: high-confidence tokens selected in the same step can rely on overlapping visual grounding, introducing visual redundancy among the committed tokens and leaving less complementary visual grounding available for later decoding. To quantify this effect, we introduce the Visual Redundancy Index (VRI), which measures visual grounding overlap among tokens committed in parallel. To control this redundancy during decoding, we propose Visual-Redundancy-Controlled Decoding (VRCD), a training-free inference-time decoding method that uses token-to-image attention to prioritize visually complementary positions. On M$^3$CoT, VRCD reduces visual redundancy and remaining-position entropy with modest runtime overhead. It improves accuracy over confidence-based unmasking on all five benchmarks, with relative gains of up to $18.8\%$ on M$^3$CoT and $6.9\%$ on MMBench in longer decoding.

Code is available at \url{https://github.com/infiniteYuanyl/VRCD}.
\end{abstract}

\section{Introduction}
\label{sec:introduction}

Diffusion-based multimodal large language models (dMLLMs) are emerging as an alternative to autoregressive multimodal generation~\citep{alayrac2022flamingo,liu2023llava,zhu2023minigpt4,bai2023qwenvl}. Instead of extending a prefix one token at a time, dMLLMs generate by iteratively decoding masked textual positions conditioned on visual input~\citep{yu2025dimple,li2025lavida,you2025lladav,yang2025mmada}. This formulation naturally enables parallel decoding: at each decoding step, the model predicts all remaining masked positions and commits only a subset of them. Parallelism can reduce the number of decoding iterations, but it also changes the nature of decoding. The central decision is no longer only which token should come next, but which masked positions should be committed together and become context for the remaining generation process.

Most existing decoding policies make this decision through token-level reliability. Confidence-based unmasking commits the top-$K$ positions with the highest predicted probabilities, while recent variants refine the process through remasking criteria, dilated schedules, revocable commitments, uncertainty-aware search, or information-gain objectives~\citep{wang2025remasking,luxembourg2025plan,hong2025wino,chen2025optimizing,yang2026infogain}. These methods improve the reliability or efficiency of diffusion decoding, but they largely preserve an implicit assumption: positions that are reliable in isolation are also suitable to commit in the same step. In multimodal decoding, this assumption can fail because textual predictions are not only statistically confident; they are also visually grounded. Thus, a parallel decoding step should be judged not only by the confidence of individual tokens, but also by whether the tokens committed together rely on complementary visual grounding.

\begin{figure}[t]
    \setlength{\abovecaptionskip}{2pt}
    \setlength{\belowcaptionskip}{-4pt}
    \noindent\begin{minipage}{\linewidth}
    \noindent\makebox[\linewidth][c]{\includegraphics[width=0.98\linewidth]{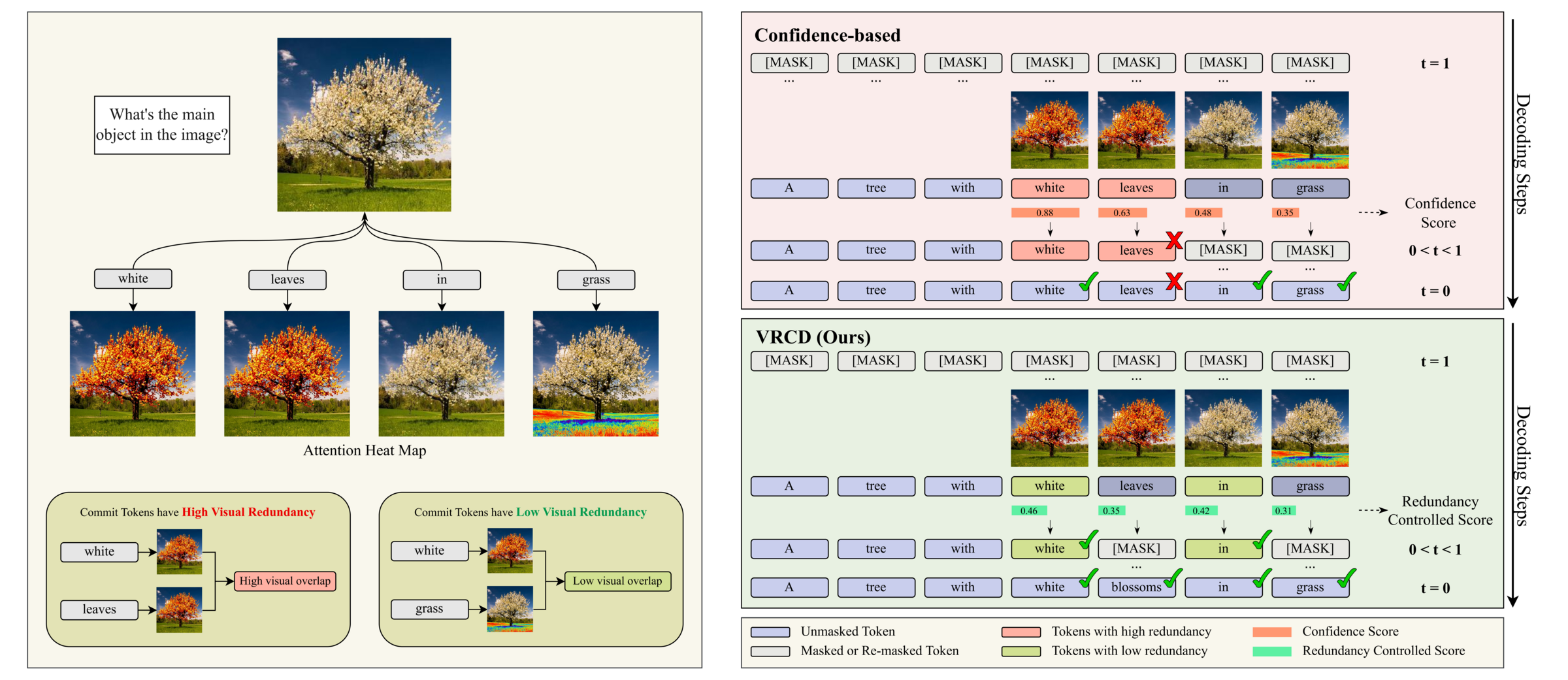}}\\[-0.25em]
    \makebox[0.44\linewidth][c]{\footnotesize \textbf{(a)} Visual redundancy}%
    \makebox[0.56\linewidth][c]{\footnotesize \textbf{(b)} Confidence-based vs. VRCD}
    \end{minipage}
    \caption{Overview of visual redundancy and VRCD. (a) Visual redundancy arises when tokens committed in the same decoding step are associated with overlapping visual grounding. (b) VRCD prioritizes lower-redundancy tokens for parallel decoding, which provides more visually complementary context for later masked positions.}
    \label{fig:vri_motivation}
    \vspace{-0.45\baselineskip}
\end{figure}

We identify an overlooked phenomenon in dMLLM parallel decoding: high-confidence textual tokens committed in the same step can rely on overlapping visual grounding. For example, several candidate tokens may attend to the same object or local image region. Although each token is reliable on its own, committing them together can repeatedly introduce similar visual grounding rather than complementary context. We call this pattern \emph{visual redundancy within the same decoding step}. It is distinct from semantic repetition in generated text and from redundancy among input visual tokens: here, redundancy refers specifically to overlap in the visual grounding of textual tokens unmasked in the same decoding step. Fig.~\ref{fig:vri_motivation}(a) illustrates this phenomenon.

This visual redundancy can shape later decoding because unmasked tokens become conditioning context for subsequent steps. Tokens with complementary visual grounding can provide more diverse cues for the remaining masked positions, whereas visually redundant tokens repeatedly ground the same region. We quantify this property with the \emph{Visual Redundancy Index} (VRI), which measures token-to-image grounding overlap among tokens committed in parallel; lower values indicate less repeated grounding within the current step.

VRI measures visual redundancy after tokens are committed, but reducing such redundancy requires a lightweight signal during decoding. Directly searching for a high-confidence, low-redundancy subset would require joint subset selection at every decoding step, which is impractical for inference-time decoding. We therefore propose \emph{Visual-Redundancy-Controlled Decoding} (VRCD), a training-free reranking method for dMLLM parallel decoding. As shown in Fig.~\ref{fig:vri_motivation}(b), at each step, VRCD builds a high-confidence candidate window, estimates pairwise visual overlap from token-to-image attention, and converts this overlap into a token-level redundancy score for confidence reweighting. This simple reweighting preserves confidence as the base reliability signal while making visually overlapping candidates less likely to be committed together. Our experiments show that VRCD lowers visual redundancy across decoding steps, improves task performance on multiple benchmarks, and retains near-baseline throughput. These results suggest that visual redundancy is a useful signal for parallel decoding in visually grounded dMLLMs.

Our contributions are summarized as follows:
\vspace{-0.45\baselineskip}
\begin{itemize}
    \setlength{\itemsep}{0pt}
    \setlength{\parskip}{0pt}
    \setlength{\parsep}{0pt}
    \setlength{\topsep}{2pt}
    \item We observe visual redundancy among textual tokens committed together in the same decoding step, an overlooked phenomenon in dMLLM parallel decoding where individually confident tokens can rely on overlapping visual grounding.
    \item We introduce the \emph{Visual Redundancy Index} (VRI), a metric that quantifies visual grounding overlap among tokens unmasked in parallel and enables systematic analysis of redundancy across decoding steps.
    \item We propose \emph{Visual-Redundancy-Controlled Decoding} (VRCD), a lightweight training-free reranking method that uses attention-derived visual overlap to prioritize high-confidence but visually complementary positions under the same decoding schedule.
\end{itemize}

\section{Related Work}
\label{sec:02_related}

\vspace{-0.6\baselineskip}
\paragraph{From autoregressive decoding to masked diffusion language models.}
Autoregressive language models generate tokens under a left-to-right factorization, whereas masked diffusion language models generate by iteratively decoding masked textual positions. Recent work also studies block-structured generation with Block Diffusion~\citep{arriola2025blockdiffusion}. More centrally for masked diffusion language modeling, Masked Discrete Diffusion for Discrete Data (MD4)~\citep{shi2024md4} and Masked Diffusion Language Models (MDLM)~\citep{sahoo2024mdlm} are particularly important: the former provides a simplified and general masked-diffusion formulation for discrete data, while the latter shows that masked diffusion can be made simple and effective for language modeling. Recent systems such as LLaDA~\citep{nie2025llada} and Dream~\citep{ye2025dream} further scale masked diffusion language modeling to large language models through modern pretraining, instruction tuning, or adaptation from autoregressive checkpoints. This shift makes decoding order a controllable choice rather than a left-to-right convention. Recent analyses study generation order~\citep{zhang2026genorder}, parallelism and generation order~\citep{zhong2026parallelism}, and dependency-guided decoding~\citep{ringel2026dependency}. Together, they show that decoding choices affect both efficiency and decoding quality.

\vspace{-0.6\baselineskip}
\paragraph{Parallel decoding for diffusion language models.}
\looseness=-1 A major advantage of diffusion language models is that they can decode multiple positions in one step, reducing the number of decoding iterations and improving inference speed. However, the standard masked-token objective factorizes over masked tokens, so parallel decoding uses marginal predictions for several positions at once rather than a joint distribution over those tokens. For dependent positions, this approximation can introduce bias; recent analyses of truly parallel decoding~\citep{li2026whyparallel} and masked diffusion parallelism~\citep{zhong2026parallelism} also report that high parallelism can reduce performance in current dLLMs. Decoding schedules therefore try to manage the tradeoff between inference speed and quality by changing which positions are unmasked. Existing methods adapt the number or layout of parallel updates, as in APD~\citep{israel2025apd} and Plan for Speed~\citep{luxembourg2025plan}. Others choose positions using Lookahead Unmasking~\citep{lee2025lookum}, information-gain sampling~\citep{yang2026infogain}, Bits2Rounds~\citep{fu2025bits2rounds}, or learned unmasking policies~\citep{jazbec2025unmasking}. These methods show that the parallel decoding strategy is important, but they still rely mainly on textual signals.

\vspace{-0.6\baselineskip}
\paragraph{Multimodal diffusion models and visual redundancy.}
Autoregressive MLLMs are important for multimodal understanding and generation and have been used in downstream tasks~\citep{wu2026vtoolr1,wang2026unimotion,wang2026skeletonllm}. Recent work also extends diffusion language modeling to multimodal understanding and generation. Dimple~\citep{yu2025dimple} introduces a discrete diffusion multimodal LLM with parallel decoding. LLaDA-V~\citep{you2025lladav} equips language diffusion models with visual instruction tuning, and LaViDa~\citep{li2025lavida} adapts large diffusion language models for multimodal understanding. MMaDA~\citep{yang2025mmada} and Lumina-DiMOO~\citep{xin2025luminadimoo} further broaden diffusion models toward unified multimodal reasoning and generation. In parallel, visual redundancy in multimodal LLMs has mostly been studied as redundancy in the visual input or in visual-token processing. Visual Context Compression~\citep{chen2024visualcontextcompression} and PyramidDrop~\citep{xing2024pyramiddrop} reduce or remove redundant visual context or tokens, while RedundancyLens~\citep{li2025redundancylens} and FALCON~\citep{zhang2025falcon} analyze or mitigate redundant visual processing in decoder-only or high-resolution MLLMs. Beyond Intermediate States~\citep{yang2025explainingvisualredundancy} explains redundant visual tokens through textual descriptions. For dMLLMs, a recent visual-token redundancy study~\citep{li2025dmllmvisualredundancy} analyzes how visual token redundancy evolves across architectures and decoding steps. Sparse-LaViDa~\citep{li2025sparselavida} accelerates multimodal diffusion sampling by truncating unnecessary masked tokens, while RedVTP~\citep{xu2025redvtp} prunes visual tokens using masked text-token attention during diffusion VLM inference. Our focus is different: rather than pruning or compressing visual tokens themselves, we study whether \emph{text tokens committed together} in a parallel decoding step are associated with overlapping visual grounding.

\section{Method}
\label{sec:03_method}

\subsection{Problem Setup}

We consider dMLLM decoding conditioned on an image $I$, a text prompt $P$, and a partially decoded token sequence $\mb{x}^{(t)}$ at step $t$. Let $C_t$ denote the set of positions that remain masked at this step. For each position $i \in C_t$, the decoder predicts a token distribution $p_i(v)=p(x_i=v\mid I, P, \mb{x}^{(t)})$, from which we obtain the most likely token $\hat{x}_i=\arg\max_v p_i(v)$ and its confidence score $c_i=p_i(\hat{x}_i)$. Let $K_t$ denote the number of tokens to commit at step $t$, as specified by the decoding schedule. Confidence-based unmasking then commits the $K_t$ highest-confidence predictions:
\begin{equation}
    S_t = \operatorname{top}_{K_t}\!\left(\{(i, c_i) \mid i \in C_t\}\right).
    \label{eq:confidence-decoding-vr}
\end{equation}
Confidence-based unmasking is efficient, but it ranks candidate tokens independently by confidence and does not account for how much visual grounding is repeated among the tokens selected for the same decoding step.

\subsection{Visual Redundancy Within Decoding Steps}

We formalize the visual redundancy illustrated in Fig.~\ref{fig:vri_motivation}(a) as an analysis metric for a realized decoding step. Given the positions $S_t$ committed at decoding step $t$, let $v_i(u)\ge 0$ denote how strongly textual token $i$ is associated with image token $u$, where $u\in\{1,\dots,N\}$ and $\sum_{u=1}^{N}v_i(u)=1$. We instantiate $v_i$ with normalized token-to-image attention as a proxy for this association~\citep{kim2026dapd}. Let $m=|S_t|$. We define the \emph{Visual Redundancy Index} (VRI) of $S_t$ as
\begin{equation}
    \operatorname{VRI}(S_t)=
    \begin{cases}
        0, & \text{if } m\le 1,\\[4pt]
        \dfrac{\sum_{u=1}^{N}\left(\sum_{i\in S_t} v_i(u)-\max_{i\in S_t} v_i(u)\right)}{m-1}, & \text{if } m\ge 2.
    \end{cases}
    \label{eq:vri-vr}
\end{equation}
With normalized $v_i$, $\operatorname{VRI}(S_t)\in[0,1]$. For each image token $u$, $\sum_{i\in S_t}v_i(u)$ is the total attention weight that the committed tokens assign to $u$, and subtracting $\max_{i\in S_t}v_i(u)$ leaves the weight from all committed tokens except the one that attends to $u$ most. A smaller VRI therefore indicates lower visual redundancy among the tokens in $S_t$. VRCD, introduced next, reduces this redundancy with a token-level redundancy-controlled score computed before the tokens are committed.

\subsection{Visual-Redundancy-Controlled Decoding}
\label{sec:vrcd}

\begin{figure}[t]
\setlength{\abovecaptionskip}{2pt}
\noindent\begin{minipage}{\linewidth}
\noindent\makebox[\linewidth][l]{\hspace*{-0.026\linewidth}\includegraphics[width=1.07\linewidth]{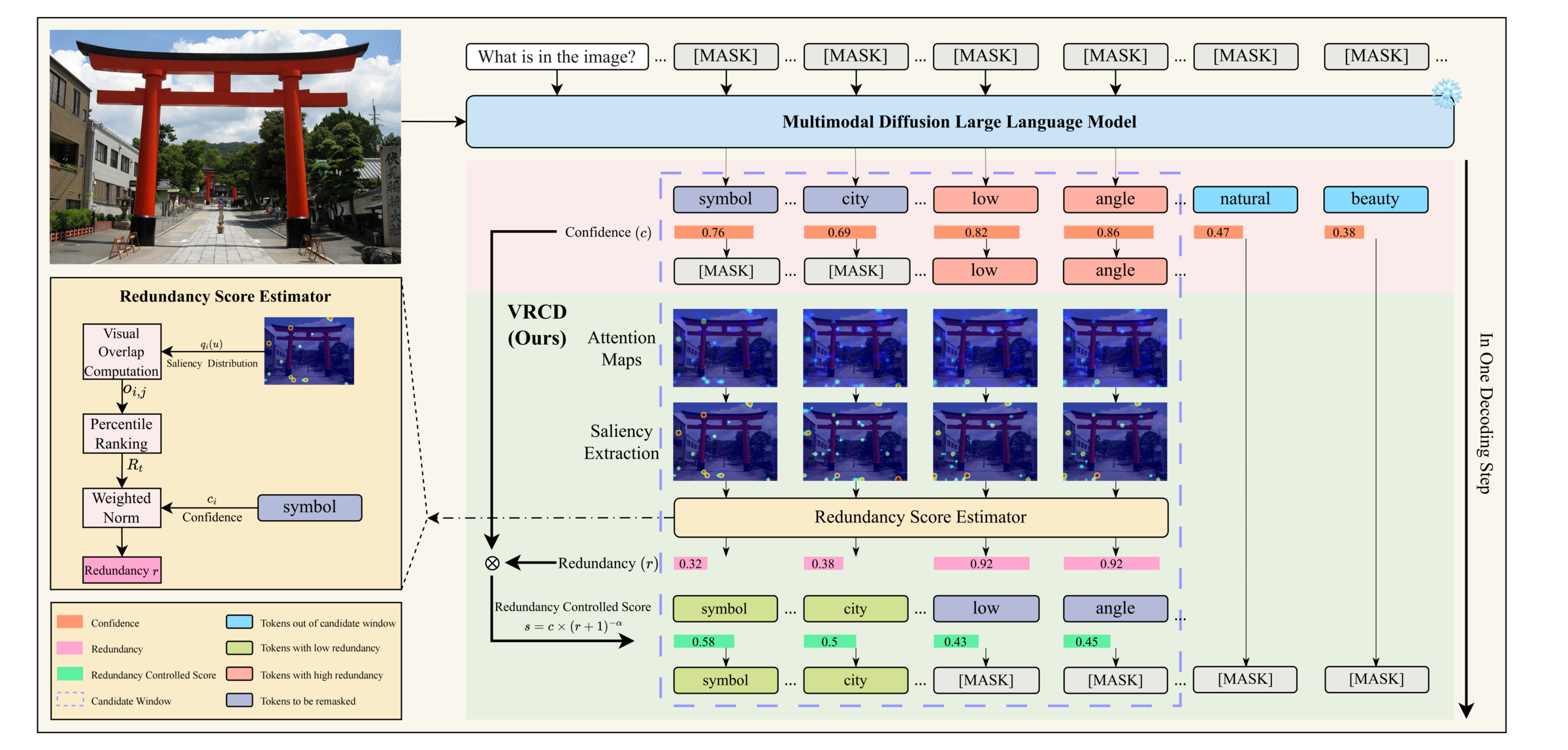}}
\end{minipage}
\caption{\textbf{Overview of VRCD:} at each decoding step, VRCD extracts visual saliency from token-to-image attention, estimates pairwise visual overlap, converts the overlaps into redundancy scores, and reweights confidence to favor visually complementary positions under the same decoding schedule. In the illustrated step, confidence-based unmasking commits the two most confident candidates, \emph{low} and \emph{angle} (red), whose visual saliency overlaps strongly; VRCD commits \emph{symbol} and \emph{city} (green), which overlap less with the other candidates. \textbf{Example output of VRCD:} The image captures a striking red torii gate, a \textcolor[RGB]{0,128,0}{\textbf{symbol}} of welcome in Shinto culture, located in a Japanese \textcolor[RGB]{0,128,0}{\textbf{city}}. The gate, with its vibrant red color, stands out against the backdrop of the cityscape. The gate is framed by lush \textcolor[RGB]{0,128,0}{\textbf{green}} \textcolor[RGB]{0,128,0}{\textbf{trees}}, adding a touch of natural beauty to the scene. Green words mark where VRCD selects different tokens from confidence-based unmasking.}
\label{fig:vrcd_overview}
\end{figure}

We propose \emph{Visual-Redundancy-Controlled Decoding} (VRCD), a decoding method that controls visual redundancy during parallel decoding. VRCD keeps confidence as the base reliability signal and adjusts the ranking only within a window of high-confidence candidates, so it retains the efficiency of confidence-based unmasking. As shown in Fig.~\ref{fig:vrcd_overview}, each decoding step builds this candidate window, extracts visual saliency from token-to-image attention, and reweights confidence by a redundancy score derived from pairwise visual overlap.

\paragraph{Candidate window.}
Estimating visual redundancy over all masked positions is unnecessary because many low-confidence tokens are unlikely to be committed at the current step. We therefore build a candidate window
\begin{equation}
    U_t = \operatorname{top}_{M_t}\!\left(\{(i, c_i) \mid i \in C_t\}\right),
    \qquad
    M_t = \min\!\left(|C_t|,\; \max\!\left(K_t, \lceil \lambda K_t \rceil\right)\right).
    \label{eq:window-size-vr}
\end{equation}
The parameter $\lambda$ controls the size of this window. This focuses redundancy estimation on positions likely to be committed at the current step and reduces the computation required by VRCD.

\paragraph{Visual saliency extraction.}
For each $i\in U_t$, we use the token-to-image attention from the final decoder layer and average it across heads. We denote the resulting normalized attention weight on image token $u$ by $a_i(u)$. To reduce attention noise, we first subtract $1/N$ from each image-token attention weight and keep only positive residual attention weights:
\begin{equation}
    \bar{a}_i(u)=\max\!\left(a_i(u)-\frac{1}{N},\,0\right).
    \label{eq:positive-residual-vr}
\end{equation}
\looseness=-1 We then renormalize the positive residual attention when at least one image token has positive residual mass:
\begin{equation}
    q_i(u)=
    \begin{cases}
        \dfrac{\bar{a}_i(u)}{\sum_{u'=1}^{N}\bar{a}_i(u')}, & \text{if } \sum_{u'=1}^{N}\bar{a}_i(u')>0,\\[8pt]
        0, & \text{otherwise.}
    \end{cases}
    \label{eq:visual-saliency-vr}
\end{equation}
Thus, $q_i$ is a normalized visual saliency distribution over image tokens after subtracting the uniform component. If all residual attention weights are non-positive, $q_i$ becomes a zero vector and the candidate is excluded from the pairwise overlap computation. This extraction highlights visual overlap, helping the redundancy score identify visual redundancy among candidates.

\paragraph{Redundancy-controlled score.}
Given $\{q_i\}_{i\in U_t}$, let \(U_t^{\mathrm{vis}}=\{i\in U_t\mid \|q_i\|_1>0\}\). For each pair \(i,j\in U_t^{\mathrm{vis}}\), we compute their pairwise visual overlap as
\(o_{ij}=\sum_{u=1}^{N}\sqrt{q_i(u)\,q_j(u)}\). Let \(\mathcal{O}_t\) collect all pairwise overlaps in the current candidate window. We convert each overlap to \(R_t(i,j)=\operatorname{PctRank}(o_{ij};\mathcal{O}_t)\), where \(\operatorname{PctRank}\) returns its percentile among \(\mathcal{O}_t\). Thus, a larger \(R_t(i,j)\) means that the two candidates share stronger visual saliency than most candidate pairs in the same step. This percentile rank makes pairwise overlaps comparable within the current candidate window. We define a token-level \emph{redundancy score} for each candidate as
\begin{equation}
    r_i =
    \begin{cases}
    0, & \text{if } i\notin U_t^{\mathrm{vis}}\text{ or }B_i=\emptyset\text{ or }Z_i=0,\\[4pt]
    \sum_{j\in B_i}\hat{c}_{j\mid i} R_t(i,j),
    & \text{otherwise}.
    \end{cases}
    \label{eq:redundancy-pressure-vr}
\end{equation}
Here \(B_i=U_t^{\mathrm{vis}}\setminus\{i\}\), \(Z_i=\sum_{\ell\in B_i}c_\ell\), and \(\hat{c}_{j\mid i}=c_j/Z_i\) when \(Z_i>0\). The confidence weight \(\hat{c}_{j\mid i}\) reflects that token \(i\) should receive a larger redundancy score when it strongly overlaps with high-confidence candidates, since these candidates are more likely to be selected for unmasking in the current decoding step. In this way, $r_i$ measures how strongly token $i$'s visual grounding overlaps with the visual grounding of other likely candidates in the current window.

\looseness=-1 We then reweight the original confidence using this redundancy score and define the redundancy-controlled score as
\begin{equation}
    s_i
    =
    c_i\,(r_i+1)^{-\alpha},
    \label{eq:commit-score-vr}
\end{equation}
where $\alpha\ge 0$ controls how strongly $r_i$ changes the confidence score. The term $r_i+1$ keeps the reweighting factor bounded and leaves $s_i=c_i$ when $r_i=0$. A large $r_i$ indicates that the visual grounding associated with token $i$ strongly overlaps with that of other likely candidates in the current window. VRCD ranks candidates by $s_i$ and selects the $K_t$ positions with the highest redundancy-controlled scores for unmasking, so candidates with stronger visual redundancy are less likely to be committed together in the same decoding step.

\subsection{Understanding VRCD from an Information-Theoretic Perspective}
\label{sec:it_view}

This subsection explains why controlling visual redundancy can improve parallel decoding. Let $X_i$ denote the token at masked position $i\in C_t$, and write $\mathcal{I}$ for mutual information to avoid confusion with the image $I$; we omit the conditioning on $I$, $P$, and $\mb{x}^{(t)}$. Parallel decoding commits the tokens in $S_t$ from their per-position predictions and thus treats them as independent. Even if each prediction equals its true marginal, the KL divergence between their product and the joint distribution of the committed tokens equals the dependence among them~\citep{benhamu2025ebsampler,zhang2026genorder}. This dependence also enters the uncertainty left for the remaining masked positions:
\begin{align}
&\mathcal{E}(S_t)=\operatorname{KL}\big(p(x_{S_t})\,\big\|\,{\textstyle\prod_{i\in S_t}}p(x_i)\big)=\sum\nolimits_{i\in S_t}H(X_i)-H(X_{S_t}),\label{eq:it-dependency-error}\\
&H\big(X_{C_t\setminus S_t}\,\big|\,X_{S_t}\big)=H(X_{C_t})-\sum\nolimits_{i\in S_t}H(X_i)+\mathcal{E}(S_t).\label{eq:it-remaining}
\end{align}
Confidence-based unmasking favors tokens with low individual entropy, which keeps the bound $\mathcal{E}(S_t)\le\sum_{i\in S_t}H(X_i)-\max_{i\in S_t}H(X_i)$ small~\citep{benhamu2025ebsampler}. The confidence score, however, is computed for each position separately and cannot account for $\mathcal{E}(S_t)$ itself.

\paragraph{Visual redundancy as dependence.}
\looseness=-1 Tokens committed in the same step are dependent when they are predicted from the same uncertain visual content. Let $V$ be a latent visual fact that the text has not yet fixed at step $t$, such as an attribute of an object. Decomposing $\mathcal{E}(S_t)$ with respect to $V$ gives
\begin{equation}
\mathcal{E}(S_t)=\mathcal{E}(S_t\mid V)+\big[\sum\nolimits_{i\in S_t}\mathcal{I}(X_i;V)-\mathcal{I}(X_{S_t};V)\big],
\label{eq:it-visual}
\end{equation}
where $\mathcal{E}(S_t\mid V)$ is the dependence that remains once $V$ is known, such as grammatical agreement. When positive, the bracketed term is visual information that the committed tokens convey more than once. It is large when several tokens state the same visual fact, such as the color of one object, and vanishes when the tokens depend on different regions whose contents are independent. Visual redundancy within a decoding step therefore contributes to $\mathcal{E}(S_t)$: the step commits the same visual fact more than once, and the committed tokens resolve less uncertainty than their individual entropies indicate.

\paragraph{Redundancy control in VRCD.}
VRCD acts on the visual overlap among tokens committed in the same step. Its redundancy-controlled score $s_i$ down-weights candidates whose extracted visual saliency overlaps with that of other likely candidates, so the tokens committed together draw on more distinct visual evidence. We take a lower pairwise visual overlap as an indication that the bracketed term in Eq.~(\ref{eq:it-visual}) is smaller. The committed tokens then share less visual information, and their dependence $\mathcal{E}(S_t)$ decreases. By Eq.~(\ref{eq:it-dependency-error}), this reduces the error of committing the tokens in parallel; by Eq.~(\ref{eq:it-remaining}), for a comparable $\sum_{i\in S_t}H(X_i)$, it also leaves less uncertainty to the remaining masked positions, so later predictions draw on more informative context. We empirically verify these effects in Sec.~\ref{sec:exp_vri_entropy} using VRI and remaining-position entropy.

\subsection{Further Discussion of Visual Redundancy}
\label{sec:discussion_vr}

Sec.~\ref{sec:it_view} reads a lower attention overlap as less shared visual information. This raises two questions: whether attention overlap reflects shared visual grounding, and when that grounding is redundant.

\looseness=-1 Token-to-image attention is a noisy proxy for visual grounding. Nearly uniform attention, common for format tokens (e.g., numeric indices or punctuation) and function words (e.g., ``and'' or ``of''), overlaps strongly while carrying little visual information. VRCD therefore computes overlap on the extracted saliency $q_i$ (Eqs.~(\ref{eq:positive-residual-vr}) and~(\ref{eq:visual-saliency-vr})), which removes this uniform component. Sec.~\ref{sec:exp_proxy} shows that the resulting pairwise overlap tracks whether the annotated image regions of two tokens coincide, and the ablation in Sec.~\ref{sec:exp_ablation} shows that the extraction makes the overlap more useful for reranking.

Shared visual grounding is redundant when several tokens state the same visual fact: the bracketed term in Eq.~(\ref{eq:it-visual}) is then large, and under the same per-step commit size $K_t$ these tokens displace candidates that would add other evidence. It is necessary when tokens grounded in the same region state different facts, such as an identity and an attribute of one object, or the parts of one time value read from a table; the bracketed term is then small despite large overlap (Appendix~\ref{app:overlap_boundary}). The redundancy score $r_i$ measures overlap and cannot separate these cases, so VRCD uses it as a soft signal: confidence remains the base score, and $\alpha$ sets how strongly $r_i$ reweights it. Larger $\alpha$ favors candidates with more complementary visual grounding at the cost of admitting candidates with lower confidence, while smaller $\alpha$ keeps the ranking closer to confidence-based unmasking; Sec.~\ref{sec:exp_ablation} examines this trade-off. The full pseudocode, implementation details, and complexity analysis are provided in Appendices~\ref{app:full_vrcd} and~\ref{app:detailed_setup}.

\section{Experiments}
\label{sec:04_experiments}

\subsection{Experimental Setup}

\paragraph{Benchmarks.}
We evaluate on five benchmarks that cover complementary multimodal abilities. \textbf{M$^3$CoT}~\citep{chen2024m3cot} tests multi-domain, multi-step multimodal reasoning with multiple-choice answers. \textbf{MMBench}~\citep{liu2024mmbench} measures general vision-language understanding across perception and reasoning skills. \textbf{SQA-IMG}~\citep{lu2022learn} denotes the image-context subset of ScienceQA, requiring models to answer multiple-choice science questions grounded in diagrams or other visual context. \textbf{DocVQA}~\citep{mathew2021docvqa} evaluates document question answering over text-rich page images, and \textbf{InfoVQA}~\citep{mathew2022infographicvqa} evaluates question answering over infographics with mixed text, layout, and visual elements. We report each benchmark's primary score, using ANLS for DocVQA and InfoVQA. All benchmarks use their official prompt formats; answer extraction and judging details are described in Appendix~\ref{app:detailed_setup}.

\paragraph{Models and baselines.}
\looseness=-1 We use \texttt{lavida-llada-reason}~\citep{li2025lavida} as the default backbone in the main experiments. All methods in the main comparison use the same model weights, prompt format, generation length limits, decoding schedules, and evaluation pipeline. In dMLLM decoding, each step predicts all currently masked positions, unmasks a subset according to the decoding schedule, and leaves the remaining positions masked for later refinement. We therefore describe local baselines as remasking policies. \textbf{Confidence} denotes confidence-based unmasking, which commits the $K_t$ positions with the highest predicted probabilities. \textbf{Margin} selects positions with the largest gap between the two most likely tokens, while \textbf{Entropy} selects positions with the lowest predictive entropy; both are local uncertainty criteria computed from the masked-diffusion predictive distribution~\citep{sahoo2024mdlm}. \textbf{IG} denotes InfoGain~\citep{yang2026infogain}, which uses information gain to choose update positions. \textbf{VRCD} is our method, which reweights confidence using visual redundancy.

\paragraph{Forward ratio and decoding length.}
We measure inference cost by the forward ratio (FR). For maximum decoding length $L$ and $T$ model forward passes, $\mathrm{FR}=T/L$. The main comparison uses $L\in\{192,384\}$ and fixed decoding schedules with $\mathrm{FR}\in\{0.125,0.25,0.5\}$, corresponding to default per-step commit sizes of $8$, $4$, and $2$ tokens.

\paragraph{Metrics and hyperparameters.}
We report VRI by decoding step and remaining-position entropy. The mechanism analysis is run on M$^3$CoT for $L\in\{192,384\}$ and $\mathrm{FR}\in\{0.25,0.5\}$; the mean-VRI table reports both lengths, and Fig.~\ref{fig:vri_entropy} shows the curves for $L=192$ and $\mathrm{FR}=0.25$. Unless otherwise stated, VRCD uses $\lambda=2.0$ in the candidate-window size definition in Eq.~(\ref{eq:window-size-vr}) and $\alpha=1.5$ in the redundancy-controlled score in Eq.~(\ref{eq:commit-score-vr}) across benchmarks and FR settings.

\begin{table}[t]
\centering
\setlength{\abovecaptionskip}{2pt}
\setlength{\belowcaptionskip}{2pt}
\scriptsize
\caption{Main comparison with decoding length 192. Best values are shown in bold.}
\label{tab:main_len192}
\setlength{\tabcolsep}{2.4pt}
\resizebox{\textwidth}{!}{%
\begin{tabular}{lccccccccccccccc}
\toprule
& \multicolumn{3}{c}{M$^3$CoT} & \multicolumn{3}{c}{MMBench} & \multicolumn{3}{c}{SQA-IMG} & \multicolumn{3}{c}{DocVQA} & \multicolumn{3}{c}{InfoVQA} \\
\cmidrule(lr){2-4} \cmidrule(lr){5-7} \cmidrule(lr){8-10} \cmidrule(lr){11-13} \cmidrule(lr){14-16}
Method / FR & 0.125 & 0.25 & 0.5 & 0.125 & 0.25 & 0.5 & 0.125 & 0.25 & 0.5 & 0.125 & 0.25 & 0.5 & 0.125 & 0.25 & 0.5 \\
\midrule
Confidence & 36.43 & 37.68 & 34.81 & 53.95 & 51.14 & 44.52 & 55.68 & 57.11 & 50.04 & 46.81 & 50.43 & 53.68 & 28.95 & 31.70 & 32.93 \\
Margin & 35.86 & 36.20 & 36.83 & 55.77 & 51.40 & 51.21 & 56.24 & 57.28 & 54.37 & 46.85 & 50.71 & 53.90 & 29.09 & 31.69 & 32.70 \\
Entropy & 32.95 & 33.48 & 27.91 & 49.68 & 41.68 & 34.86 & 51.69 & 48.24 & 32.04 & 46.72 & 49.95 & 53.00 & 29.18 & 31.30 & 32.69 \\
IG & 36.66 & 37.39 & 34.56 & 54.06 & 51.16 & 45.68 & 55.53 & 57.31 & 50.67 & 46.81 & 50.70 & 53.80 & 29.02 & 31.69 & \textbf{33.26} \\
VRCD & \textbf{37.94} & \textbf{40.38} & \textbf{39.60} & \textbf{58.79} & \textbf{59.37} & \textbf{57.50} & \textbf{60.09} & \textbf{63.38} & \textbf{58.64} & \textbf{49.27} & \textbf{52.30} & \textbf{54.34} & \textbf{29.93} & \textbf{32.61} & 33.06 \\
\bottomrule
\end{tabular}
}
\par\vspace{3pt}
\caption{Main comparison with decoding length 384. Best values are shown in bold.}
\label{tab:main_len384}
\setlength{\tabcolsep}{2.4pt}
\resizebox{\textwidth}{!}{%
\begin{tabular}{lccccccccccccccc}
\toprule
& \multicolumn{3}{c}{M$^3$CoT} & \multicolumn{3}{c}{MMBench} & \multicolumn{3}{c}{SQA-IMG} & \multicolumn{3}{c}{DocVQA} & \multicolumn{3}{c}{InfoVQA} \\
\cmidrule(lr){2-4} \cmidrule(lr){5-7} \cmidrule(lr){8-10} \cmidrule(lr){11-13} \cmidrule(lr){14-16}
Method / FR & 0.125 & 0.25 & 0.5 & 0.125 & 0.25 & 0.5 & 0.125 & 0.25 & 0.5 & 0.125 & 0.25 & 0.5 & 0.125 & 0.25 & 0.5 \\
\midrule
Confidence & 28.82 & 37.16 & 38.35 & 57.29 & 63.04 & 66.91 & 53.93 & 62.96 & 66.06 & 46.45 & 50.41 & 53.74 & 29.01 & 31.83 & 32.81 \\
Margin & 34.11 & 38.06 & 38.47 & 60.64 & 66.43 & 68.48 & \textbf{63.38} & \textbf{64.98} & 65.54 & 46.54 & 50.57 & 53.74 & 29.03 & 31.87 & 32.96 \\
Entropy & 26.33 & 34.57 & 35.86 & 50.30 & 57.48 & 69.02 & 42.44 & 54.47 & 63.56 & 46.39 & 49.93 & 52.93 & 28.89 & 31.57 & 32.79 \\
IG & 28.72 & 37.26 & 39.23 & 57.54 & 63.35 & 67.73 & 54.12 & 62.30 & 66.57 & 46.48 & 50.51 & 54.05 & 28.94 & 31.81 & \textbf{33.37} \\
VRCD & \textbf{34.24} & \textbf{39.82} & \textbf{41.99} & \textbf{61.24} & \textbf{66.73} & \textbf{69.87} & 56.62 & 64.49 & \textbf{68.43} & \textbf{49.22} & \textbf{52.24} & \textbf{54.33} & \textbf{30.37} & \textbf{32.55} & 32.94 \\
\bottomrule
\end{tabular}
}
\end{table}

\subsection{Main Comparison}
\label{sec:exp_main}

We vary FR to test the setting where dMLLMs commit multiple tokens in parallel: lower FR values use larger per-step commit sizes, making the choice of which positions are committed together more consequential. VRCD improves over confidence-based unmasking at every FR setting and at both decoding lengths (Tabs.~\ref{tab:main_len192} and~\ref{tab:main_len384}).

\looseness=-1 The gains are largest on M$^3$CoT, MMBench, and SQA-IMG, whose answers often involve longer multimodal reasoning. Tokens committed early in this reasoning become context for later steps, so the choice of which tokens are committed together carries over to the rest of decoding. Gains on DocVQA and InfoVQA are smaller, since their prompts tend to elicit short answers and leave fewer later positions for this effect. VRCD also improves over the MMaDA baseline in most settings (Appendix~\ref{app:mmada_backbone}).

\noindent
\begin{minipage}[t]{0.715\linewidth}
\vspace{0pt}
\textbf{Throughput.} Tab.~\ref{tab:throughput_inline} reports throughput on full M$^3$CoT with decoding length $L=192$ and $\mathrm{FR}=0.5$. VRCD adds about $1.5\%$ runtime overhead over confidence-based unmasking. In contrast, IG~\citep{yang2026infogain} incurs a much larger slowdown under the same setting. These results show that VRCD remains a lightweight decoding method. Complexity and measurement details are deferred to Appendix~\ref{app:full_vrcd}.
\end{minipage}\hspace{0.005\linewidth}%
\begin{minipage}[t]{0.28\linewidth}
\vspace{0pt}
\centering
\small
\refstepcounter{table}\label{tab:throughput_inline}
Table~\thetable: Throughput on M$^3$CoT ($L=192$, $\mathrm{FR}=0.5$).\par\vspace{2pt}
\begin{tabular}{@{}lc@{}}
\toprule
Method & tokens/s \\
\midrule
LaViDa & 46.982 \\
LaViDa+IG & 24.884 \\
LaViDa+VRCD & 46.264 \\
\bottomrule
\end{tabular}
\end{minipage}

\subsection{Redundancy and Predictive Certainty}
\label{sec:exp_vri_entropy}

\looseness=-1 We study the two effects described in Sec.~\ref{sec:it_view} on M$^3$CoT: whether tokens committed in the same step share less visual grounding, measured by VRI, and whether the remaining masked positions are predicted with less uncertainty, measured by remaining-position entropy.

\begin{wraptable}[7]{r}{0.305\linewidth}
\vspace{-1.0\baselineskip}
\centering
\small
\refstepcounter{table}\label{tab:vri_inline}
Table~\thetable: Mean VRI on M$^3$CoT.\par\vspace{2pt}
\setlength{\tabcolsep}{2.5pt}
\resizebox{\linewidth}{!}{%
\begin{tabular}{@{}ccccc@{}}
\toprule
$L$ & FR & LaViDa & IG & VRCD \\
\midrule
192 & 0.25 & 0.752 & 0.749 & \textbf{0.712} \\
192 & 0.5 & 0.685 & 0.679 & \textbf{0.588} \\
384 & 0.25 & 0.784 & 0.783 & \textbf{0.743} \\
384 & 0.5 & 0.720 & 0.718 & \textbf{0.623} \\
\bottomrule
\end{tabular}
}
\vspace{-0.7\baselineskip}
\end{wraptable}

\begin{figure}[t]
    \centering
    \setlength{\abovecaptionskip}{2pt}
    \setlength{\belowcaptionskip}{1pt}
    \includegraphics[width=0.96\linewidth]{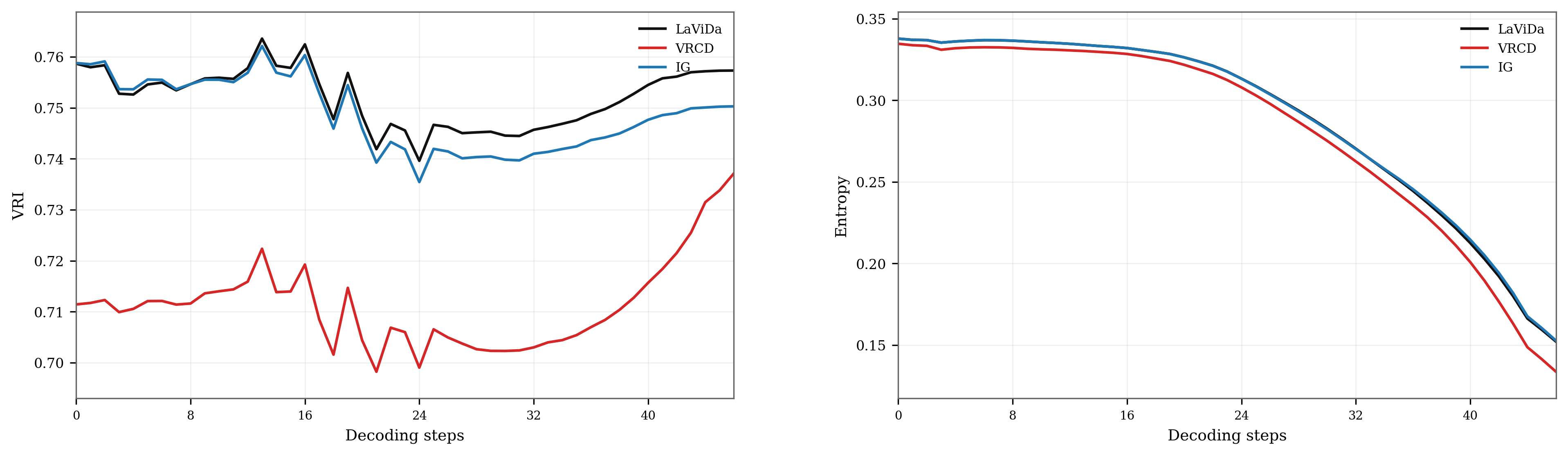}
    \par\vspace{-0.55em}
    \noindent\makebox[\linewidth][l]{%
    \begin{minipage}[t]{0.49\linewidth}
        \centering
        \makebox[\linewidth][c]{\footnotesize (a) VRI}
    \end{minipage}
    \hfill
    \begin{minipage}[t]{0.49\linewidth}
        \centering
        \makebox[\linewidth][c]{\footnotesize (b) Remaining-position entropy}
    \end{minipage}%
    }
    \vspace{-2pt}
    \caption{M$^3$CoT VRI and entropy curves ($L=192$, $\mathrm{FR}=0.25$).}
    \label{fig:vri_entropy}
    \vspace{0pt}

    \noindent\makebox[\linewidth][l]{%
    \begin{minipage}[t]{0.49\linewidth}
        \centering
        \includegraphics[width=0.95\linewidth]{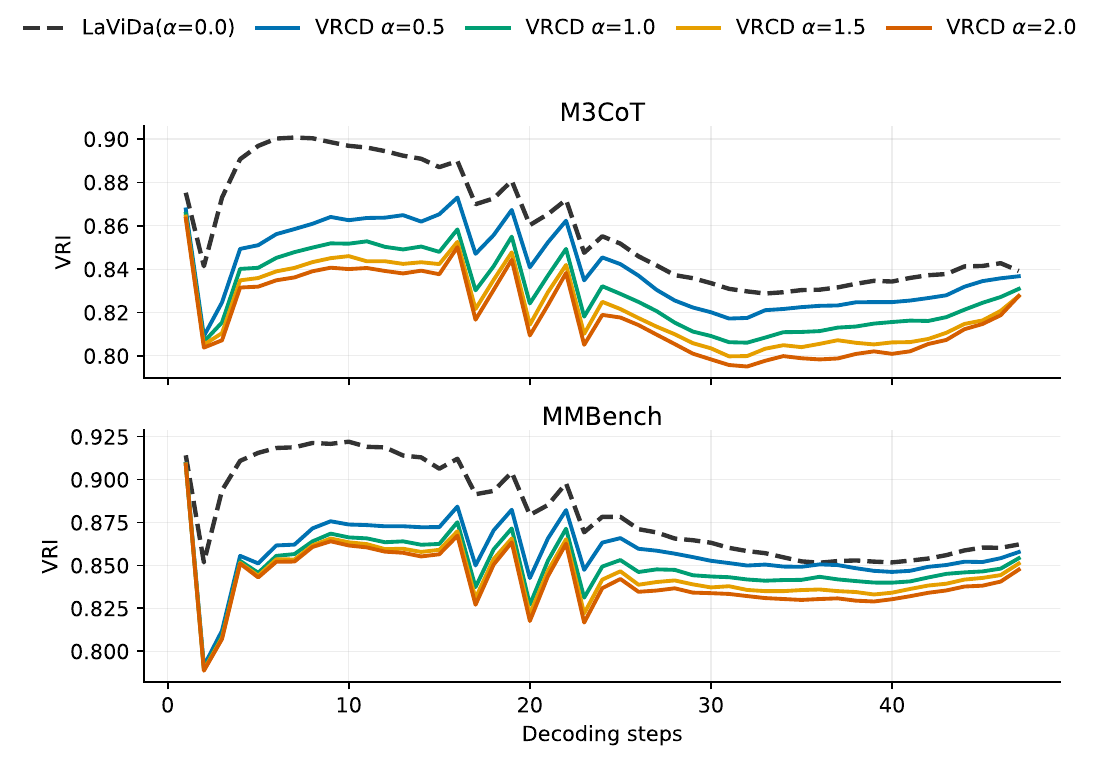}
        \par\vspace{-0.55em}
        \makebox[\linewidth][c]{\footnotesize (a) VRI}
    \end{minipage}
    \hfill
    \begin{minipage}[t]{0.49\linewidth}
        \centering
        \includegraphics[width=0.95\linewidth]{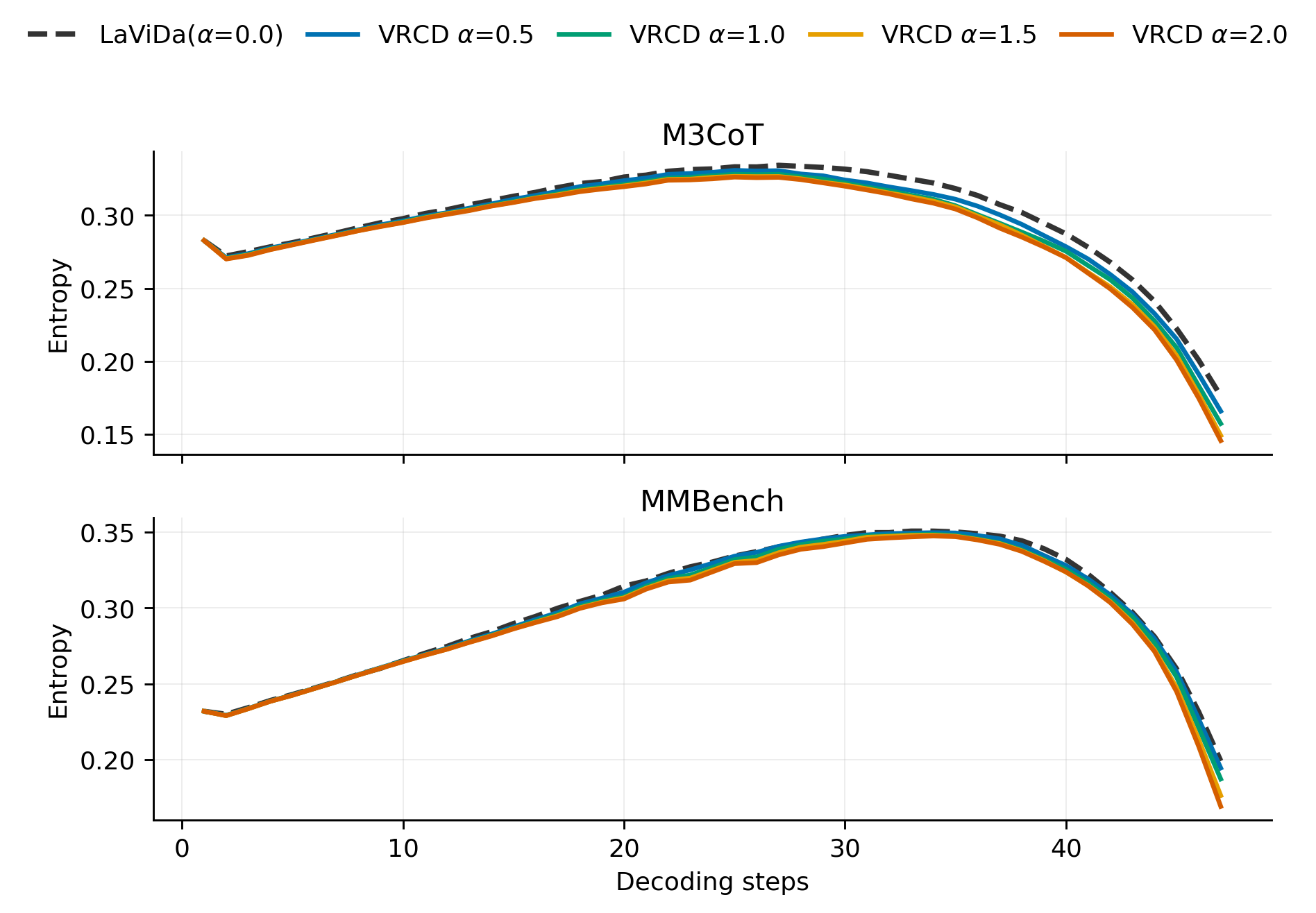}
        \par\vspace{-0.55em}
        \makebox[\linewidth][c]{\footnotesize (b) Remaining-position entropy}
    \end{minipage}%
    }
    \vspace{-2pt}
    \caption{Effect of $\alpha$ on M$^3$CoT and MMBench ($L=384$, $\mathrm{FR}=0.125$).}
    \label{fig:alpha_vri_entropy}
\end{figure}

\paragraph{Reducing VRI during decoding.}
In all four settings of Tab.~\ref{tab:vri_inline}, VRCD has a lower mean VRI than both LaViDa and IG~\citep{yang2026infogain}. Fig.~\ref{fig:vri_entropy}(a) tracks VRI by decoding step at $L=192$ and $\mathrm{FR}=0.25$. VRCD stays below LaViDa throughout decoding, while IG, which also changes the selection criterion, stays close to LaViDa. Reweighting confidence by visual overlap thus changes which tokens are committed together, and the tokens committed in the same step repeat less visual grounding.

\paragraph{Increasing later predictive certainty.}
\looseness=-1 VRCD also leaves less uncertainty for the remaining masked positions. In Fig.~\ref{fig:vri_entropy}(b), its remaining-position entropy stays below that of LaViDa and IG throughout decoding, and the gap widens in later steps. Together with the lower VRI, this is in line with the second effect in Sec.~\ref{sec:it_view}: when tokens committed in the same step share less visual information, later predictions are made with greater certainty.

\subsection{Ablations}
\label{sec:exp_ablation}

We ablate the three design choices that determine VRCD's reranking behavior: the exponent $\alpha$ in the redundancy-controlled score in Eq.~(\ref{eq:commit-score-vr}), redundancy score aggregation, and visual saliency extraction. The $\alpha$ sweep uses M$^3$CoT and MMBench at decoding length $L=384$ and $\mathrm{FR}=0.125$, the setting with the largest per-step commit size. The component ablations for redundancy score aggregation and visual saliency extraction use the same FR setting and report both $L=192$ and $L=384$.

\begin{wraptable}[8]{r}{0.315\linewidth}
\vspace{-0.65\baselineskip}
\centering
\begingroup
\small
\refstepcounter{table}\label{tab:alpha_inline}Table~\thetable: Effect of $\alpha$.\par\vspace{2pt}
\setlength{\tabcolsep}{3pt}
\begin{tabular}{@{}ccc@{}}
\toprule
\(\alpha\) & M$^3$CoT & MMBench \\
\midrule
0.0 & 28.82 & 57.29 \\
0.5 & 29.97 (+1.15) & 59.03 (+1.74) \\
1.0 & 31.54 (+2.72) & 59.79 (+2.50) \\
1.5 & \textbf{34.24 (+5.42)} & 61.24 (+3.95) \\
2.0 & 33.45 (+4.63) & \textbf{62.69 (+5.40)} \\
\bottomrule
\end{tabular}
\endgroup
\vspace{-0.5\baselineskip}
\end{wraptable}

\noindent\textbf{Effect of $\alpha$.} We evaluate $\alpha \in \{0.0,0.5,1.0,1.5,2.0\}$ under the default configuration. Here $\alpha=0.0$ corresponds to confidence-based unmasking, while larger $\alpha$ gives the redundancy score $r_i$ more influence in Eq.~(\ref{eq:commit-score-vr}). Tab.~\ref{tab:alpha_inline} reports accuracy for each value; parenthesized values denote gains over $\alpha=0.0$. On M$^3$CoT, $\alpha=1.5$ is best in this grid, while on MMBench $\alpha=2.0$ is best. We therefore keep $\alpha=1.5$ as a fixed moderate setting across benchmarks and FR settings. Fig.~\ref{fig:alpha_vri_entropy} reports the corresponding VRI and remaining-position entropy curves.

\vspace{0.4\baselineskip}

\noindent
\begin{minipage}[t]{0.455\linewidth}
\vspace{0pt}
\textbf{Redundancy score aggregation.} To test whether $r_i$ should account for how likely an overlapping neighbor is to be committed in the same decoding step, we compare uniform average aggregation with the default confidence-weighted aggregation. The average variant averages $R_t(i,j)$ over $j\in B_i$, while the weighted variant uses the normalized confidence weights in Eq.~(\ref{eq:redundancy-pressure-vr}). Tab.~\ref{tab:redundancy_score_aggregation_inline} reports each entry as $L=192/384$ and shows that confidence-weighted aggregation performs better on both benchmarks.
\end{minipage}\hspace{0.015\linewidth}%
\begin{minipage}[t]{0.525\linewidth}
\vspace{0pt}
\centering
{\small\refstepcounter{table}\label{tab:redundancy_score_aggregation_inline}Table~\thetable: Redundancy score aggregation.\par}\vspace{2pt}
\begingroup
\footnotesize
\setlength{\tabcolsep}{0pt}
\begin{tabular*}{0.98\linewidth}{@{\extracolsep{\fill}}lcc@{}}
\toprule
Agg. & M$^3$CoT (192/384) & MMBench (192/384) \\
\midrule
Average & 37.35/32.17 & 58.10/60.63 \\
Weighted & \textbf{37.94/34.24} & \textbf{58.79/61.24} \\
\bottomrule
\end{tabular*}
\endgroup

\vspace{0.55\baselineskip}

{\small\refstepcounter{table}\label{tab:saliency_extraction_inline}Table~\thetable: Visual saliency extraction (VSE).\par}\vspace{2pt}
\begingroup
\footnotesize
\setlength{\tabcolsep}{0pt}
\begin{tabular*}{0.98\linewidth}{@{\extracolsep{\fill}}lcc@{}}
\toprule
Variant & M$^3$CoT (192/384) & MMBench (192/384) \\
\midrule
Confidence & 36.43/28.82 & 53.95/57.29 \\
VRCD (w/o VSE) & 37.59/34.13 & 58.25/61.00 \\
VRCD & \textbf{37.94/34.24} & \textbf{58.79/61.24} \\
\bottomrule
\end{tabular*}
\endgroup
\end{minipage}

\vspace{0.4\baselineskip}

\looseness=-1 \noindent\textbf{Visual saliency extraction (VSE).} We compare VRCD with VRCD (w/o VSE), a variant that computes pairwise visual overlap from raw token-to-image attention without this module. Tab.~\ref{tab:saliency_extraction_inline} reports both decoding lengths. Confidence and VRCD values are taken from Tabs.~\ref{tab:main_len192} and~\ref{tab:main_len384}. VRCD (w/o VSE) is consistently below VRCD, indicating that VSE makes the overlap estimate more useful in these settings.

\subsection{Attention Overlap as a Visual-Grounding Proxy}
\label{sec:exp_proxy}

\looseness=-1 The redundancy score of VRCD is useful only if attention overlap reflects shared visual grounding (Sec.~\ref{sec:discussion_vr}). A proxy for shared grounding should have two properties: tokens grounded in the same image region should overlap more than tokens grounded in different regions, and this difference should come from the image content. We test both on Flickr30K Entities~\citep{plummer2015flickr30k}, which links caption phrases to annotated image boxes. In each of 307 test images, we choose two phrases whose boxes overlap little, mask two tokens in each phrase, and let the model predict them with the other caption words visible; the boxes are used only for evaluation. This yields same-phrase pairs, which refer to one annotated region, and cross-phrase pairs, which refer to two largely separate regions. For each pair, we compute the pairwise visual overlap $o_{ij}$ as VRCD does (Sec.~\ref{sec:vrcd}) and keep the 510 pairs in which both tokens are predicted correctly.

\looseness=-1 Attention overlap separates the two kinds of pairs. The mean $o_{ij}$ of same-phrase pairs is 0.148 higher than that of cross-phrase pairs, and $o_{ij}$ is positively correlated with the overlap between the annotated regions of the two tokens (Spearman $\rho=0.484$); the 95\% bootstrap intervals of both lie above zero. To check that this separation comes from the image, we replace each image with an unrelated one and keep the caption and mask positions unchanged. The gap then falls from 0.148 to 0.073, so the image content accounts for about half of the separation; the correlation $\rho$ drops as well. Pairwise attention overlap is therefore an effective proxy for shared visual grounding: it distinguishes tokens grounded in the same region from tokens grounded in different regions, and the distinction depends on the image. This is the property that the redundancy score of VRCD relies on. Appendix~\ref{app:proxy_validation} provides more details.

\section{Conclusion}

We studied visual redundancy within decoding steps in diffusion-based multimodal large language models. Our analysis shows that confidence-based unmasking can commit several individually confident tokens whose attention-derived visual saliency overlaps within the same step. To quantify this behavior, we introduced the Visual Redundancy Index (VRI), and we proposed VRCD, a lightweight inference-time decoding method that estimates pairwise visual overlap from token-to-image attention and reweights confidence under the same decoding schedule.

\looseness=-1 Empirically, VRCD reduces visual redundancy across decoding steps and is accompanied by lower entropy over remaining masked positions, consistent with higher predictive certainty for later predictions. It also improves decoding quality while retaining near-baseline throughput. Together, these experiments demonstrate that visual redundancy serves as a useful signal for deciding which positions are committed in parallel. More broadly, this work shifts the decoding question from whether individual predictions are confident to whether a parallel decoding step introduces complementary visual grounding for the remaining generation process. Future work can explore adaptive multimodal decoding methods that use visual redundancy to guide both which positions are committed together and the per-step commit size within the decoding schedule.

\begin{ack}
This work was supported by the National Natural Science Foundation of China (NSFC) under Grant No.~62306277.
We declare no competing interests.
\end{ack}

\bibliographystyle{plainnat}
\bibliography{main}

\clearpage
\appendix
\raggedbottom
\makeatletter
\setlength{\@fptop}{0pt}
\setlength{\@fpsep}{\floatsep}
\setlength{\@fpbot}{0pt plus 1fil}
\makeatother

\section{Appendix}

\subsection{Full VRCD Procedure and Implementation Details}
\label{app:full_vrcd}

This section provides the full pseudocode of VRCD and specifies the implementation details that are summarized in the main paper.

\begin{algorithm}[H]
\caption{Visual-Redundancy-Controlled Decoding (VRCD) at decoding step $t$}
\label{alg:vrcd-vr}
\begin{algorithmic}[1]
\Require masked positions $C_t$, per-step commit size $K_t$, predicted tokens $\{\hat{x}_i\}_{i\in C_t}$, confidences $\{c_i\}_{i\in C_t}$, window multiplier $\lambda$, exponent $\alpha$
\State $M_t \gets \min\!\left(|C_t|,\max\!\left(K_t,\lceil \lambda K_t\rceil\right)\right)$
\State $U_t \gets$ the $M_t$ positions in $C_t$ with the largest confidences $c_i$
\For{each $i \in U_t$} \Comment{extract visual saliency}
    \State Extract normalized head-averaged final-layer attention $a_i(u)$ over image tokens
    \State Compute $q_i(u)$ according to Eq.~(\ref{eq:visual-saliency-vr})
\EndFor
\State $U_t^{\mathrm{vis}} \gets \{i\in U_t \mid \|q_i\|_1>0\}$ \Comment{candidates with nonzero visual saliency}
\For{each $i\in U_t$}
    \State $R_t(i,i)\gets 0$
\EndFor
\For{each pair $(i,j)$ with $i,j\in U_t^{\mathrm{vis}}$ and $i<j$} \Comment{compute all pairwise overlaps}
    \State $o_{ij} \gets \sum_{u=1}^{N}\sqrt{q_i(u)\,q_j(u)}$
\EndFor
\State $\mathcal{O}_t \gets \{o_{ab}\mid a,b\in U_t^{\mathrm{vis}},\ a<b\}$
\For{each pair $(i,j)$ with $i,j\in U_t^{\mathrm{vis}}$ and $i<j$} \Comment{rank overlaps after all pairs are available}
    \State $R_t(i,j) \gets \operatorname{PctRank}(o_{ij};\mathcal{O}_t)$
    \State $R_t(j,i)\gets R_t(i,j)$
\EndFor
\For{each $i \in U_t$} \Comment{compute normalized confidence-weighted redundancy scores}
    \State $B_i \gets U_t^{\mathrm{vis}}\setminus\{i\}$ and $Z_i\gets\sum_{\ell\in B_i}c_\ell$
    \State $r_i\gets 0$ if $i\notin U_t^{\mathrm{vis}}$, $B_i=\emptyset$, or $Z_i=0$; otherwise $r_i\gets \sum_{j\in B_i}(c_j/Z_i)R_t(i,j)$
    \State $s_i \gets c_i(r_i+1)^{-\alpha}$
\EndFor
\State Commit the $K_t$ predicted tokens in $U_t$ with the largest redundancy-controlled scores $s_i$
\end{algorithmic}
\end{algorithm}

\paragraph{Implementation details.}
In all experiments, the candidate window is rebuilt at every decoding step using the current masked positions and the current $K_t$. Candidates with no positive residual attention have $q_i=\mathbf{0}$. They are excluded from the pairwise overlaps and from every $B_i$, receive $r_i=0$ (Eq.~(\ref{eq:redundancy-pressure-vr})), and remain eligible for selection with $s_i=c_i$. The percentile rank $\operatorname{PctRank}(o_{ij};\mathcal{O}_t)$ is the rank of $o_{ij}$ in $\mathcal{O}_t$, counted from 0 and divided by $|\mathcal{O}_t|-1$. Tied values share their average rank, and all ranks are 0 when $\mathcal{O}_t$ has fewer than two distinct values. VRCD does not add extra model-evaluation forward passes; the additional computation is limited to attention post-processing, pairwise overlap calculation, and confidence reweighting inside the current candidate window. The main experiments were run on NVIDIA L40 GPUs, and the other experiments were run on NVIDIA GeForce RTX 4090 GPUs.

\paragraph{Attention-map processing.}
\label{app:attention_processing}
At each decoding step, VRCD takes the final-layer attention from each position $i\in U_t$ to the $N$ image tokens, averages it over heads, and normalizes it to sum to one, which gives $a_i(u)$. This attention comes from the forward pass that produces the confidences. The number $N$ depends on the aspect ratio of the image (Tab.~\ref{tab:appendix_base_setup}). VRCD subtracts $1/N$ from each $a_i(u)$, keeps the positive part, and renormalizes it to obtain $q_i$ (Eqs.~(\ref{eq:positive-residual-vr}) and~(\ref{eq:visual-saliency-vr})). The pairwise visual overlap $o_{ij}=\sum_{u=1}^{N}\sqrt{q_i(u)\,q_j(u)}$ is the Bhattacharyya coefficient between two extracted visual saliency distributions. It is 1 when they coincide and 0 when they share no image token. VRI applies Eq.~(\ref{eq:vri-vr}) to $v_i=a_i$ without the $1/N$ subtraction. The redundancy score $r_i$ and the proxy validation (Appendix~\ref{app:proxy_validation}) use $q_i$ and $o_{ij}$. The proxy validation restricts $q_i$ to the spatial image tokens, where the boxes are defined, and renormalizes it.

\paragraph{Computational cost and throughput.}
Let $M_t=|U_t|$ and let $N$ be the number of image tokens. After the model forward pass produces token-to-image attention, visual saliency extraction costs $O(M_tN)$. Computing pairwise overlaps over the current candidate window costs $O(M_t^2N)$, and ranking the overlaps and computing $r_i$ add $O(M_t^2\log M_t)$. Since \(M_t=O(K_t)\) by construction, the additional post-processing scales with the candidate window rather than with all masked positions. Tab.~\ref{tab:throughput_inline} reports decoding throughput, the number of tokens committed per second of decoding time, which includes this post-processing.

\subsection{Detailed Experimental Configuration}
\label{app:detailed_setup}

The default model checkpoint is \texttt{lavida-llada-reason}, i.e., LaViDa-Reason with LLaDA-8B as the diffusion language backbone~\citep{li2025lavida,nie2025llada}. The vision encoder is SigLIP-400M within LaViDa's any-resolution pipeline, which encodes three to five views per image (Tab.~\ref{tab:appendix_base_setup}).

\begin{table}[H]
\centering
\small
\caption{Fixed implementation configuration used across experiments.}
\label{tab:appendix_base_setup}
\setlength{\tabcolsep}{3pt}
\begin{tabular}{p{0.30\linewidth}p{0.60\linewidth}}
\toprule
Setting & Value \\
\midrule
Model checkpoint & \texttt{lavida-llada-reason} (LaViDa-Reason) \\
Diffusion language backbone & LLaDA-8B~\citep{nie2025llada} \\
Vision encoder & SigLIP-400M~\citep{li2025lavida} \\
Image preprocessing & any-resolution: a global view plus two to four $384\times384$ crops, each bilinearly downsampled to $14\times14$ tokens \\
Visual sequence & depends on the aspect ratio; 1008 positions for a square image (980 spatial tokens, 28 row separators) \\
Attention source for VRCD & final decoder layer, averaged over attention heads \\
Evaluation implementation & official prompts and scoring code where available; \texttt{qwen3:8b} for final-answer extraction on M$^3$CoT, MMBench, and SQA-IMG; GPT-5.6 as judge for the open-ended benchmarks \\
\bottomrule
\end{tabular}
\end{table}

All methods in the main comparison share the same backbone, prompt format, decoding schedule, per-step commit-size schedule, generation limits, and evaluation pipeline. We report $L=192$ and $L=384$. Unless ablated, VRCD fixes $\lambda=2.0$ in Eq.~(\ref{eq:window-size-vr}) and $\alpha=1.5$ in Eq.~(\ref{eq:commit-score-vr}) across all datasets and FR settings, without per-benchmark tuning. The supplementary $\lambda$ sweep uses M$^3$CoT at $L=192$ and $\mathrm{FR}\in\{0.125,0.25,0.5\}$. The $\alpha$ sweep uses M$^3$CoT and MMBench at $L=384$ and $\mathrm{FR}=0.125$, and component ablations use both datasets at both lengths under the same FR setting.

\paragraph{Local remasking scores and remaining-position entropy.}
For every decoding step $t$ and masked position $i\in C_t$, all local remasking policies use the same predictive distribution $p_i(v)=p(x_i=v\mid I,P,\mathbf{x}^{(t)})$ over the vocabulary $\mathcal{V}$. Let $v_i^{(1)}$ and $v_i^{(2)}$ be the most likely and second most likely tokens under $p_i$. The confidence score is
\begin{equation}
    c_i=\max_{v\in\mathcal{V}}p_i(v)=p_i(v_i^{(1)}),
    \label{eq:appendix-confidence-score}
\end{equation}
and confidence-based unmasking selects the $K_t$ positions with the largest $c_i$. The margin score is
\begin{equation}
    m_i=p_i(v_i^{(1)})-p_i(v_i^{(2)}),
    \label{eq:appendix-margin-score}
\end{equation}
and the margin baseline selects the $K_t$ positions with the largest $m_i$. The entropy baseline uses normalized predictive entropy
\begin{equation}
    h_i=-\frac{1}{\log|\mathcal{V}|}\sum_{v\in\mathcal{V}}p_i(v)\log p_i(v),
    \label{eq:appendix-entropy-score}
\end{equation}
and selects the $K_t$ positions with the smallest $h_i$. The normalization in Eq.~(\ref{eq:appendix-entropy-score}) only rescales entropy values and does not change the ranking within a decoding step.

For the remaining-position entropy curves, after a method chooses which positions to unmask at step $t$, let $C_{n,t}^{\mathrm{rem}}$ be the positions that remain masked for sample $n$. We compute this quantity on non-terminal decoding steps, so $|C_{n,t}^{\mathrm{rem}}|>0$. The per-sample remaining-position entropy is
\begin{equation}
    e_{n,t}=
    \dfrac{1}{|C_{n,t}^{\mathrm{rem}}|}
    \sum_{i\in C_{n,t}^{\mathrm{rem}}}h_{n,t,i},
    \label{eq:appendix-remaining-entropy-example}
\end{equation}
where $h_{n,t,i}$ is Eq.~(\ref{eq:appendix-entropy-score}) for position $i$ in sample $n$ at step $t$. The plotted value at decoding step $t$ is the average of $e_{n,t}$ over samples at that step.

\paragraph{Benchmark prompting and answer extraction.}
All benchmarks use official prompt formats, and we use official scorers when available. For M$^3$CoT, MMBench, and SQA-IMG, we use \texttt{qwen3:8b} only to map generated text to a final multiple-choice option before scoring, with instructions not to infer a new answer from the question, image, or outside knowledge. GPT-5.6 judges the open-ended benchmarks in Appendix~\ref{app:open_ended}.

\subsection{Limitations, Societal Impact, and Assets}
\label{app:responsible_reporting}

\paragraph{Limitations.}
VRCD changes which masked positions are committed together under a given decoding schedule, but it does not adapt the decoding schedule or the number of positions committed at each step. If the schedule commits too many or too few positions for a task, VRCD can only rerank the available candidates within that schedule. Dynamically choosing how many tokens to commit could require searching over candidate subsets, whose cost can grow combinatorially with the candidate window size. We leave adaptive commit-size selection to future work.
VRCD also relies on the quality of token-to-image attention: when attention is diffuse or misaligned with the relevant image regions, attention overlap is a weaker proxy for shared visual grounding, and our test of this proxy against human region annotations checks correspondence rather than causal attribution (Appendix~\ref{app:proxy_validation}). In addition, the gains are small in some open-ended settings with stochastic sampling, and VRCD with $\alpha=1.5$ scores below confidence-based unmasking in one deterministic and one sampled setting (Appendix~\ref{app:open_ended}).

\paragraph{Ethics and broader impacts.}
This work proposes an inference-time decoding method for existing dMLLMs and does not introduce a new dataset, train a new model, or collect new human-subject data. A positive impact is that VRCD can improve multimodal decoding quality without additional model training and with near-baseline throughput in our measurement. A potential negative impact is that better multimodal decoding can also improve systems used in settings where visual or document understanding errors affect users, including sensitive document analysis. The method should therefore be evaluated in the target domain before use in high-stakes settings, and it should be paired with the safety practices and usage restrictions of the underlying model and datasets.

\paragraph{Existing assets and licenses.}
Tab.~\ref{tab:appendix_assets_licenses} summarizes the public license or terms information we found for the main assets used in the experiments. We use all datasets only for evaluation and do not redistribute benchmark data.

\begin{table}[tp]
\centering
\scriptsize
\caption{Existing assets used in the experiments and public license or terms information. ``Not specified'' means that we did not find an explicit license field on the public page used for that asset during manuscript preparation.}
\label{tab:appendix_assets_licenses}
\setlength{\tabcolsep}{2.5pt}
\begin{tabular}{>{\raggedright\arraybackslash}p{0.21\linewidth}>{\raggedright\arraybackslash}p{0.29\linewidth}>{\raggedright\arraybackslash}p{0.44\linewidth}}
\toprule
Asset & Use in this paper & Public license or terms noted \\
\midrule
\texttt{lavida-llada-reason} & Backbone checkpoint~\citep{li2025lavida} & License not specified on the public Hugging Face checkpoint page. \\
LaViDa codebase~\citep{li2025lavida} & Base of the released VRCD code & Apache-2.0 license on the public LaViDa GitHub repository. \\
LLaDA-8B~\citep{nie2025llada} & Diffusion language backbone used by LaViDa & MIT license on the public LLaDA-8B model page. \\
SigLIP-400M & Vision encoder used by LaViDa & Apache-2.0 license on the public SigLIP model page. \\
MMaDA-8B-MixCoT~\citep{yang2025mmada} & Backbone for Appendix~\ref{app:mmada_backbone} & MIT license on the public MMaDA-8B-MixCoT model page. \\
\texttt{qwen3:8b} & Final-answer extraction for M$^3$CoT, MMBench, and SQA-IMG & Apache-2.0 license on the public Qwen3-8B model page. \\
GPT-5.6 & Automatic judge for the open-ended benchmarks & Proprietary OpenAI model without public weights; accessed through OpenAI's Codex service and used under OpenAI's terms of use. \\
M$^3$CoT~\citep{chen2024m3cot} & Multimodal reasoning benchmark & MIT license on the public M$^3$CoT dataset page. \\
MMBench~\citep{liu2024mmbench} & General vision-language benchmark & Apache-2.0 license on the public MMBench repository. \\
ScienceQA~\citep{lu2022learn} & Multimodal science question answering benchmark & The public ScienceQA repository states that the code is MIT licensed and that the dataset is licensed under CC BY-NC-SA 4.0. \\
DocVQA~\citep{mathew2021docvqa} & Document question answering benchmark & The public LMMs-Lab mirror lists Apache-2.0; the official DocVQA page states that task images are from the UCSF Industry Documents Library. \\
InfoVQA~\citep{mathew2022infographicvqa} & Infographic question answering benchmark & The official InfographicVQA page describes challenge access and states that images are collected from the Internet; no separate license field is listed there. \\
MM-Vet~\citep{yu2024mmvet} & Open-ended benchmark (Appendix~\ref{app:open_ended}) & The public MM-Vet repository states that the code is under the Apache-2.0 license and the dataset under the CC BY-NC 4.0 license. \\
LLaVA-Bench-in-the-Wild~\citep{liu2023llava} & Open-ended benchmark (Appendix~\ref{app:open_ended}) & Apache-2.0 license on the public Hugging Face dataset page. \\
Flickr30K Entities~\citep{plummer2015flickr30k} & Proxy validation (Appendix~\ref{app:proxy_validation}) & The public repository states that use of the images must abide by the Flickr Terms of Use and that the images are provided for non-commercial research and educational purposes; no separate license is listed for the annotations. \\
\bottomrule
\end{tabular}
\end{table}

\subsection{Additional Backbone Evaluation on MMaDA}
\label{app:mmada_backbone}
\suppressfloats[t]

We apply VRCD to MMaDA-8B-MixCoT~\citep{yang2025mmada} and compare it with the MMaDA baseline under the same prompt format, decoding schedule, per-step commit-size schedule, generation limits, and evaluation pipeline. VRCD uses the same $\alpha=1.5$ and $\lambda=2.0$ as with LaViDa. Prompts follow the released MMaDA MixCoT evaluation setting, which adds the official chain-of-thought instruction to multiple-choice and VQA prompts. The evaluation covers M$^3$CoT, MMBench, SQA-IMG, DocVQA, and InfoVQA, with $L\in\{192,384\}$ and $\mathrm{FR}\in\{0.125,0.25,0.5\}$. VRCD improves over the MMaDA baseline in most settings (Tab.~\ref{tab:appendix_mmada_backbone}).

\begin{table}[tb]
\centering
\footnotesize
\caption{Additional MMaDA backbone evaluation. MMaDA denotes the native confidence-based unmasking baseline; MMaDA+VRCD applies the same visual-redundancy-controlled decoding criterion under the matched MMaDA setting. Values are task scores, with ANLS used for DocVQA and InfoVQA. Best values in each matched setting are bolded.}
\label{tab:appendix_mmada_backbone}
\setlength{\tabcolsep}{2pt}
\begin{tabular}{lccccccccccccccc}
\toprule
& \multicolumn{3}{c}{M$^3$CoT} & \multicolumn{3}{c}{MMBench} & \multicolumn{3}{c}{SQA-IMG} & \multicolumn{3}{c}{DocVQA} & \multicolumn{3}{c}{InfoVQA} \\
\cmidrule(lr){2-4} \cmidrule(lr){5-7} \cmidrule(lr){8-10} \cmidrule(lr){11-13} \cmidrule(lr){14-16}
Method / FR & 0.125 & 0.25 & 0.5 & 0.125 & 0.25 & 0.5 & 0.125 & 0.25 & 0.5 & 0.125 & 0.25 & 0.5 & 0.125 & 0.25 & 0.5 \\
\midrule
\multicolumn{16}{c}{$L=192$} \\
\midrule
MMaDA & 24.42 & 27.61 & 28.42 & 52.09 & 53.45 & 50.66 & 44.75 & 43.86 & \textbf{49.22} & 8.49 & 10.18 & 11.19 & 7.65 & 10.08 & 12.40 \\
MMaDA+VRCD & \textbf{26.01} & \textbf{29.09} & \textbf{29.31} & \textbf{53.45} & \textbf{54.84} & \textbf{52.87} & \textbf{47.51} & \textbf{45.10} & 46.79 & \textbf{9.09} & \textbf{11.38} & \textbf{11.88} & \textbf{8.40} & \textbf{12.05} & \textbf{13.33} \\
\midrule
\multicolumn{16}{c}{$L=384$} \\
\midrule
MMaDA & 24.68 & 25.50 & 26.85 & 50.90 & 50.82 & 44.65 & 47.08 & 43.10 & 45.51 & 9.27 & 11.23 & 11.64 & 8.43 & 10.87 & 11.80 \\
MMaDA+VRCD & \textbf{26.19} & \textbf{28.24} & \textbf{28.03} & \textbf{52.81} & \textbf{53.52} & \textbf{51.32} & \textbf{48.29} & \textbf{46.47} & \textbf{47.43} & \textbf{9.58} & \textbf{11.74} & \textbf{12.38} & \textbf{8.86} & \textbf{12.05} & \textbf{14.20} \\
\bottomrule
\end{tabular}
\end{table}

\subsection{Candidate Window Size and Position Changes}
\label{app:window_position_change}

\paragraph{Candidate window size.}
We vary the window multiplier $\lambda$ on M$^3$CoT at decoding length $L=192$. The grid uses $\lambda\in\{1.5,2.0,2.5,3.0\}$ with $\mathrm{FR}\in\{0.125,0.25,0.5\}$. We exclude $\lambda=1.0$, which would leave the candidate window identical to the positions selected by confidence-based unmasking. Tab.~\ref{tab:appendix_candidate_window_accuracy} reports M$^3$CoT accuracy. In this grid, $\lambda=2.5$ gives the best value at all three FR settings.

\begin{table}[H]
\centering
\small
\caption{M$^3$CoT accuracy under different $\lambda$ values ($L=192$).}
\label{tab:appendix_candidate_window_accuracy}
\begin{tabular}{cccccc}
\toprule
FR & Confidence & $\lambda=1.5$ & $\lambda=2.0$ & $\lambda=2.5$ & $\lambda=3.0$ \\
\midrule
0.125 & 36.43 & 37.17 & 37.94 & \textbf{38.09} & 37.59 \\
0.25  & 37.68 & 38.05 & 40.38 & \textbf{40.81} & 40.38 \\
0.5   & 34.81 & 33.49 & 39.60 & \textbf{40.51} & 39.69 \\
\bottomrule
\end{tabular}
\end{table}

\paragraph{Position-change rate.}
We measure how often VRCD changes which positions are committed in parallel relative to confidence-based unmasking. For sample $n$ and decoding step $t$ of a VRCD run, let $S_{n,t}^{\mathrm{VRCD}}$ be the positions that VRCD commits and $S_{n,t}^{\mathrm{conf}}$ the top-$K_{n,t}$ positions by confidence in the same decoding state. We aggregate over non-terminal decoding steps $\mathcal{T}=\{(n,t):0<K_{n,t}<|C_{n,t}|\}$, where more candidate positions are available than can be committed at that step.
\begin{align}
D_{n,t} &= K_{n,t} - \bigl|S_{n,t}^{\mathrm{conf}} \cap S_{n,t}^{\mathrm{VRCD}}\bigr|,
\label{eq:changed-positions}\\
\bar{D} &= \frac{\sum_{(n,t)\in\mathcal{T}} D_{n,t}}{|\mathcal{T}|},
\qquad
\eta = \frac{\sum_{(n,t)\in\mathcal{T}} D_{n,t}}{\sum_{(n,t)\in\mathcal{T}} K_{n,t}}.
\label{eq:changed-position-rate}
\end{align}
These quantities are computed on M$^3$CoT and MMBench for $L\in\{192,384\}$ and $\mathrm{FR}\in\{0.125,0.25\}$, as shown in Tab.~\ref{tab:appendix_position_change_rate}. Here $D_{n,t}$ counts the positions that VRCD commits outside the top-$K_{n,t}$ positions by confidence, $\bar{D}$ averages this count over the analyzed steps, and the position-change rate $\eta$ is the fraction of committed positions that VRCD changes. The rate measures how strongly VRCD alters the position selection.

\begin{table}[htb]
\centering
\small
\caption{Position-change statistics for VRCD. $\bar{D}$ is the average position-change count per decoding step, and $\eta$ is the position-change rate.}
\label{tab:appendix_position_change_rate}
\begin{tabular}{lcccc}
\toprule
Dataset & $L$ & FR & $\bar{D}$ & $\eta$ \\
\midrule
M$^3$CoT & 192 & 0.125 & 1.31 & 0.164 \\
M$^3$CoT & 192 & 0.25 & 1.03 & 0.258 \\
M$^3$CoT & 384 & 0.125 & 1.41 & 0.177 \\
M$^3$CoT & 384 & 0.25 & 0.96 & 0.240 \\
MMBench & 192 & 0.125 & 1.09 & 0.137 \\
MMBench & 192 & 0.25 & 0.79 & 0.197 \\
MMBench & 384 & 0.125 & 1.28 & 0.160 \\
MMBench & 384 & 0.25 & 0.89 & 0.222 \\
\bottomrule
\end{tabular}
\end{table}

\subsection{Validation of Attention Overlap as a Visual-Grounding Proxy}
\label{app:proxy_validation}

VRCD uses the pairwise visual overlap $o_{ij}$ between extracted visual saliency distributions as a proxy for shared visual grounding. This section details the test in Sec.~\ref{sec:exp_proxy}.

\paragraph{Setup.}
(i)~\emph{Data.} From the 1{,}000 test images of Flickr30K Entities~\citep{plummer2015flickr30k}, we select one caption per image with two phrases that pass length, scene, and box-size filters and whose boxes barely overlap (Bhattacharyya coefficient of the box masks at most 0.10); 307 images qualify.
(ii)~\emph{Masking.} We mask the first and the last content token of each phrase, keep the rest of the caption visible, and prompt the model to complete the description from the image. One forward pass predicts the four masked tokens, and we take their final-layer attention as in Appendix~\ref{app:attention_processing}.
(iii)~\emph{Pairs.} We keep the same-phrase and cross-phrase pairs whose two tokens are both predicted correctly, so that the boxes of each phrase annotate the tokens the model predicts; this leaves 510 pairs from 216 of the 307 images.
(iv)~\emph{Measures.} For each pair, we compute the attention overlap $o_{ij}$ and the annotated-region overlap, i.e., the same Bhattacharyya coefficient applied to the boxes of the two phrases mapped onto the image tokens (1 for same-phrase pairs). We report the gap in mean $o_{ij}$ between same-phrase and cross-phrase pairs and the Spearman correlation $\rho$ between the two overlaps.
(v)~\emph{Control.} We replace each image with another of the 307 images, resized to the original size so that the image-token layout is unchanged, and recompute $o_{ij}$ for the same pairs.

\begin{table}[H]
\centering
\small
\caption{Attention overlap on Flickr30K Entities (510 token pairs, final decoder layer). Brackets give 95\% bootstrap intervals over images (10{,}000 resamples).}
\label{tab:appendix_proxy_validation}
\setlength{\tabcolsep}{6pt}
\begin{tabular}{lcc}
\toprule
Image given to the model & Same-phrase minus cross-phrase gap & Spearman $\rho$ \\
\midrule
Original image & 0.148 [0.127, 0.169] & 0.484 [0.405, 0.559] \\
Unrelated image & 0.073 [0.056, 0.091] & 0.343 [0.251, 0.434] \\
\midrule
Original minus unrelated & 0.075 [0.057, 0.092] & 0.141 [0.064, 0.220] \\
\bottomrule
\end{tabular}
\end{table}

\paragraph{Results.}
In Tab.~\ref{tab:appendix_proxy_validation}, the 95\% intervals of the same-phrase minus cross-phrase gap, of $\rho$, and of their reductions under an unrelated image all lie above zero. The correlation $\rho$ also partly reflects how far apart the two tokens are in the caption and how confident their predictions are. The test evaluates pairwise overlap, the quantity that VRCD uses.

\subsection{Token Sampling, Open-Ended Generation, and Decoder Layer}
\label{app:additional_experiments}

The main experiments rest on three defaults: every method commits the most likely token at each selected position, the benchmarks score multiple-choice or short answers, and VRCD takes visual saliency from the final decoder layer. This section relaxes each default in turn. VRCD keeps its gains over confidence-based unmasking when the committed tokens are sampled (Appendix~\ref{app:stochastic_sampling}) and when answers are free-form (Appendix~\ref{app:open_ended}), and the final decoder layer gives more useful visual saliency than the two alternatives we compare (Appendix~\ref{app:layer_ablation}). Unless stated otherwise, all runs use \texttt{lavida-llada-reason}, $L=192$, $\lambda=2.0$, $\alpha=1.5$, and the decoding schedules of Sec.~\ref{sec:04_experiments}; sampled runs draw each committed token by Top-$k$ ($k=50$) or Top-$p$ ($p=0.9$) sampling at temperature 1.0.

\subsubsection{Stochastic Token Sampling}
\label{app:stochastic_sampling}

VRCD decides which positions are committed together, while the token at each committed position comes from the model's prediction. To check that the gains do not depend on committing the most likely token, we sample the committed tokens and compare VRCD with confidence-based unmasking on M$^3$CoT. We sample one output per question, score it three times with the answer-extraction pipeline of the main experiments, and report the mean. VRCD scores higher than confidence-based unmasking at every FR under both samplers (Tab.~\ref{tab:app_sampling}). Choosing which positions to commit together thus remains effective when the committed tokens are sampled.

\begin{table}[H]
\centering
\small
\caption{M$^3$CoT accuracy with stochastic token sampling ($L=192$, $\alpha=1.5$). Each entry is the mean of three scoring runs on one sampled output per question. Bold: better result in each column for each sampler.}
\label{tab:app_sampling}
\setlength{\tabcolsep}{5pt}
\begin{tabular}{@{}llccc@{}}
\toprule
Sampler & Method / FR & 0.125 & 0.25 & 0.5 \\
\midrule
Top-$k$ ($k=50$) & Confidence & 30.57 & 35.13 & 33.07 \\
& VRCD & \textbf{33.72} & \textbf{38.48} & \textbf{36.86} \\
\midrule
Top-$p$ ($p=0.9$) & Confidence & 33.10 & 35.29 & 34.01 \\
& VRCD & \textbf{34.77} & \textbf{37.86} & \textbf{37.83} \\
\bottomrule
\end{tabular}
\end{table}

\subsubsection{Open-Ended Generation}
\label{app:open_ended}

To see whether the gains extend to free-form answers, we evaluate MM-Vet~\citep{yu2024mmvet} (218 questions) and LLaVA-Bench-in-the-Wild~\citep{liu2023llava} (LLaVA-Wild, 60 questions), judged by GPT-5.6 with the official prompts and rubrics. MM-Vet reports an overall score on a 0--100 scale, and LLaVA-Wild reports a relative score, the mean rating of the generated answers as a percentage of that of the reference answers. We compare VRCD with confidence-based unmasking and Margin under deterministic decoding and under the two samplers of Appendix~\ref{app:stochastic_sampling}. Deterministic entries average at least three judge runs of one output, and sampled entries average three seeds with three judge runs each. Besides the default $\alpha=1.5$, we report $\alpha=2.0$, the best value on MMBench in Tab.~\ref{tab:alpha_inline}.

\begin{table}[tbp]
\centering
\small
\caption{Open-ended generation on MM-Vet and LLaVA-Wild ($L=192$). Entries are GPT-5.6 judge scores (higher is better): the overall score on MM-Vet and the relative score on LLaVA-Wild. Det.: deterministic decoding. Bold: best score in each column, per benchmark.}
\label{tab:app_open_ended}
\setlength{\tabcolsep}{3.5pt}
\begin{tabular}{@{}lccccccccc@{}}
\toprule
& \multicolumn{3}{c}{Det.} & \multicolumn{3}{c}{Top-$k$ ($k=50$)} & \multicolumn{3}{c}{Top-$p$ ($p=0.9$)} \\
\cmidrule(lr){2-4} \cmidrule(lr){5-7} \cmidrule(lr){8-10}
Method / FR & 0.125 & 0.25 & 0.5 & 0.125 & 0.25 & 0.5 & 0.125 & 0.25 & 0.5 \\
\midrule
\multicolumn{10}{@{}l}{\textit{MM-Vet}} \\
Confidence & 24.62 & 23.20 & 24.79 & 22.02 & 19.91 & 21.64 & 20.76 & 20.22 & 22.37 \\
Margin & 24.67 & 24.13 & 22.64 & 20.65 & 22.53 & \textbf{23.51} & 22.61 & 22.58 & \textbf{23.88} \\
VRCD ($\alpha=1.5$) & 26.93 & 25.46 & 24.58 & 21.51 & \textbf{23.04} & 22.32 & 22.78 & 23.10 & 23.03 \\
VRCD ($\alpha=2.0$) & \textbf{28.67} & \textbf{26.30} & \textbf{25.02} & \textbf{22.48} & 22.23 & 22.18 & \textbf{22.98} & \textbf{23.14} & 23.33 \\
\midrule
\multicolumn{10}{@{}l}{\textit{LLaVA-Wild}} \\
Confidence & 16.39 & 18.84 & 15.97 & 15.91 & 16.79 & 16.91 & 16.68 & 16.23 & 17.72 \\
Margin & \textbf{19.67} & 17.39 & 17.68 & 14.68 & 14.75 & \textbf{19.42} & 15.84 & 16.42 & \textbf{19.95} \\
VRCD ($\alpha=1.5$) & 19.20 & 19.11 & \textbf{18.79} & \textbf{17.93} & 17.74 & 18.64 & 16.76 & \textbf{18.79} & 18.50 \\
VRCD ($\alpha=2.0$) & 18.36 & \textbf{19.50} & 18.49 & 17.57 & \textbf{19.00} & 19.20 & \textbf{17.60} & 18.10 & 18.93 \\
\bottomrule
\end{tabular}
\end{table}

\paragraph{Results.}
VRCD improves over confidence-based unmasking in open-ended generation under both deterministic decoding and sampling (Tab.~\ref{tab:app_open_ended}): with $\alpha=2.0$ it scores higher in every setting, and with $\alpha=1.5$ in all but two. VRCD also outperforms Margin in most settings, although Margin scores highest in the sampled settings at $\mathrm{FR}=0.5$. Because VRCD changes only which positions are committed together, these results show that reranking by visual overlap also benefits free-form generation.

\subsubsection{Decoder Layer for Visual Saliency}
\label{app:layer_ablation}

VRCD takes visual saliency from the token-to-image attention of the final decoder layer (Sec.~\ref{sec:vrcd}), the same layer whose output gives the confidences that VRCD reweights. To check whether another layer gives more useful visual saliency for the redundancy score, we replace the final layer with one of two alternatives: a single layer at about two-thirds of the depth, or the attention averaged uniformly over the last third of the layers. The final layer gives the highest M$^3$CoT accuracy at all three FR values (Tab.~\ref{tab:app_layer}); the last-third average comes close, and the two-thirds-depth layer is the lowest at every FR. Attention closer to the output thus gives a more useful estimate of visual overlap, and we keep the final layer as the default.

\begin{table}[H]
\centering
\small
\caption{VRCD accuracy on M$^3$CoT ($L=192$, $\alpha=1.5$) with visual saliency from three choices of decoder layers. Bold: best accuracy at each FR. The final-layer column repeats the VRCD results on M$^3$CoT in Tab.~\ref{tab:main_len192}.}
\label{tab:app_layer}
\setlength{\tabcolsep}{6pt}
\begin{tabular}{@{}cccc@{}}
\toprule
FR & Final layer & Two-thirds depth & Last-third average \\
\midrule
0.125 & \textbf{37.94} & 32.96 & 37.76 \\
0.25 & \textbf{40.38} & 35.36 & 39.72 \\
0.5 & \textbf{39.60} & 34.60 & 38.75 \\
\bottomrule
\end{tabular}
\end{table}

\subsection{When Is Visual Overlap Redundant?}
\label{app:overlap_boundary}

Visual overlap plays two roles in parallel decoding. When tokens committed together repeat the same visual fact, the overlap is redundant: it enlarges the bracketed term in Eq.~(\ref{eq:it-visual}) and, under the same per-step commit size $K_t$, displaces candidates that would add new evidence. When tokens grounded in the same region state different facts, such as an identity and an attribute, or an action and a relation, the overlap reflects shared grounding that the answer needs (Sec.~\ref{sec:discussion_vr}). The redundancy score sees only the overlap, which is why VRCD reweights confidence softly and leaves the strength to $\alpha$. Below, we show where each role appears, first in the $\alpha$ ablation and then in individual questions.

\looseness=-1 The $\alpha$ ablation traces how the balance shifts as the control grows stronger (Tab.~\ref{tab:alpha_inline} and Fig.~\ref{fig:alpha_vri_entropy}). On both benchmarks, a larger $\alpha$ lowers VRI. On MMBench, accuracy rises with $\alpha$ over the whole range, so removing repeated visual evidence keeps paying off. On M$^3$CoT, accuracy peaks at $\alpha=1.5$ and dips at $\alpha=2.0$ while VRI keeps falling, which indicates that the strongest setting begins to suppress shared grounding that the answers rely on. VRCD therefore keeps confidence as the base score and uses a moderate default: $\alpha=1.5$ is the best setting on M$^3$CoT and still yields a clear gain on MMBench.

The case studies in Appendix~\ref{app:case_studies} show the same balance within single questions. In Fig.~\ref{fig:case_positive}, the answer depends on two separate labels in the image, and lowering overlap lets VRCD state the evidence from both labels before it commits the answer. In Fig.~\ref{fig:case_failure}, the answer is a single table cell whose tokens share the same grounding, and lowering their overlap splits one reading across steps. Together, the ablation and the cases show where lower overlap stops helping, and $\alpha$ controls how close VRCD comes to that point.

\subsection{Case Studies}
\label{app:case_studies}

\begin{figure}[!htb]
    \centering
    \includegraphics[width=0.90\linewidth]{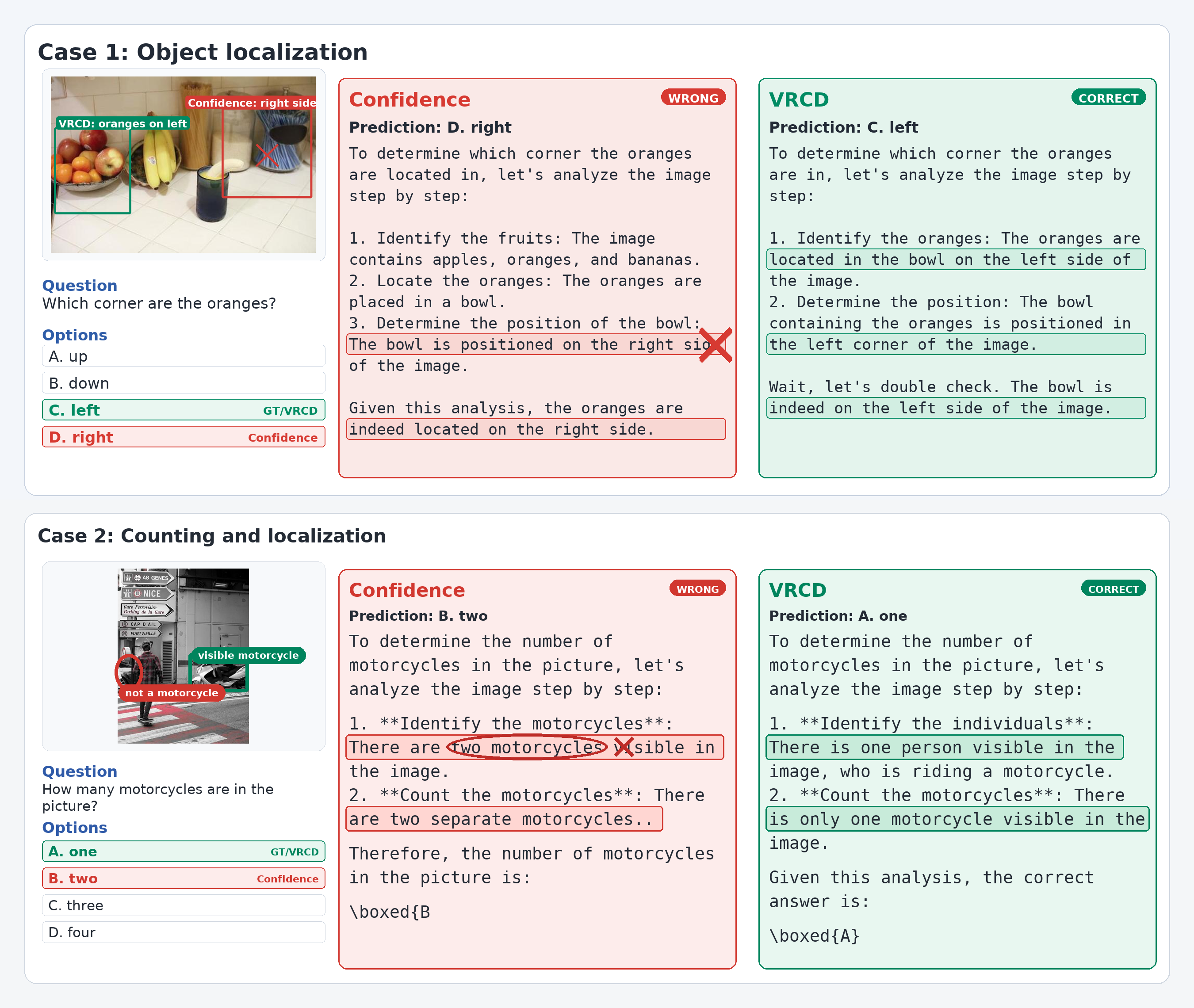}
    \caption{Qualitative MMBench examples. The left column shows the image, question, and choices; the red and green panels show output excerpts. In the object-localization example, confidence-based unmasking assigns the fruit bowl to the right side, while VRCD keeps the oranges on the left. In the counting example, confidence-based unmasking treats the partial car as another motorcycle, while VRCD keeps the count at the single visible motorcycle.}
    \label{fig:mmbench_case_study}
\end{figure}

\begin{figure}[tp]
    \centering
    \includegraphics[width=\linewidth]{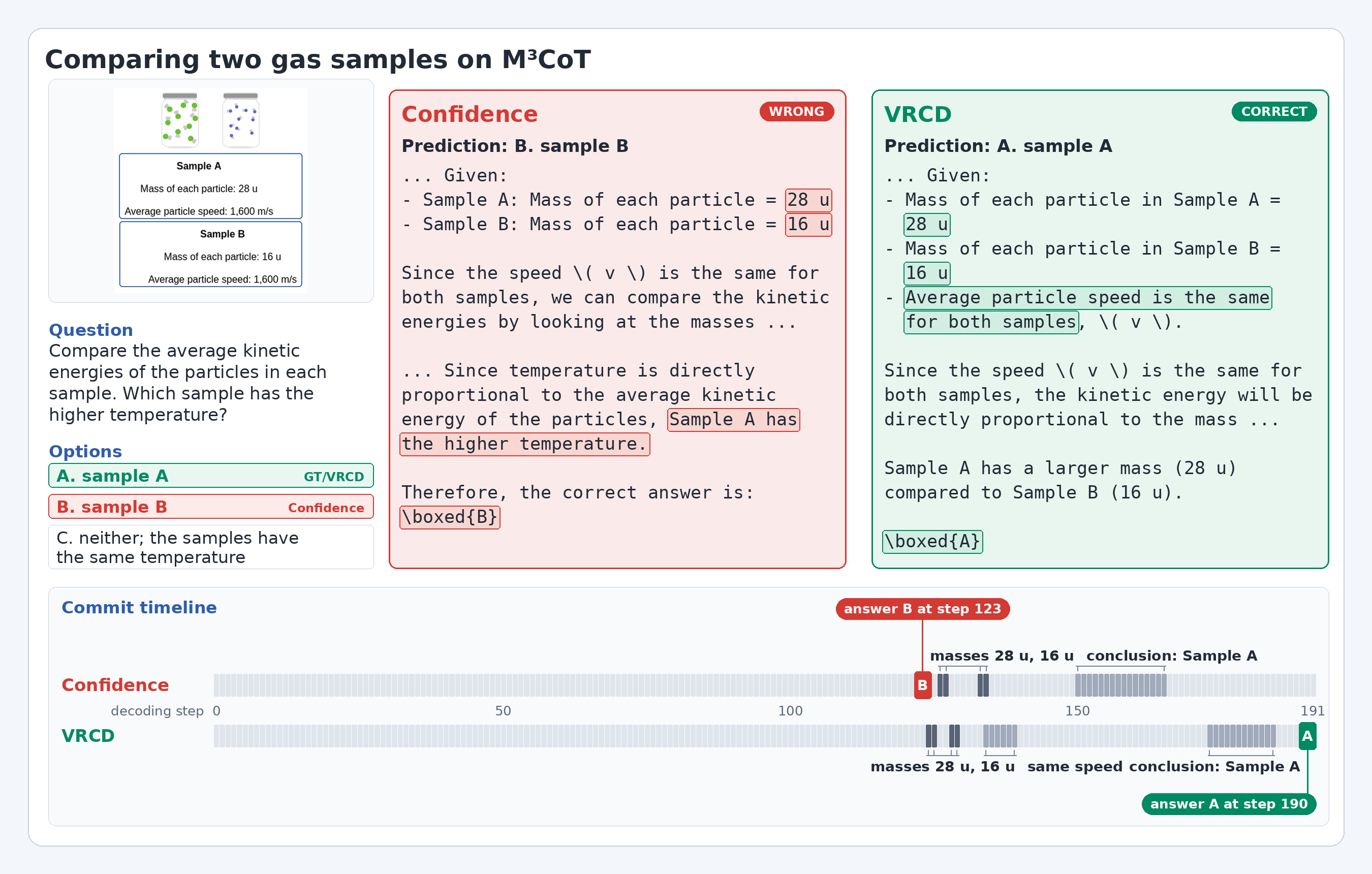}
    \caption{Gas-sample question from M$^3$CoT ($L=384$, $\mathrm{FR}=0.5$). Left: image with the two sample labels enlarged, question, and options. Right: output excerpts with key phrases boxed. Bottom: the steps at which each method commits its answer, the two masses, and its conclusion; the VRCD timeline also marks the statement that the two speeds are equal. Confidence-based unmasking commits B at step 123, before either mass; VRCD commits A at step 190, after both masses and the equal speeds.}
    \label{fig:case_positive}
\end{figure}

\begin{figure}[tp]
    \centering
    \includegraphics[width=\linewidth]{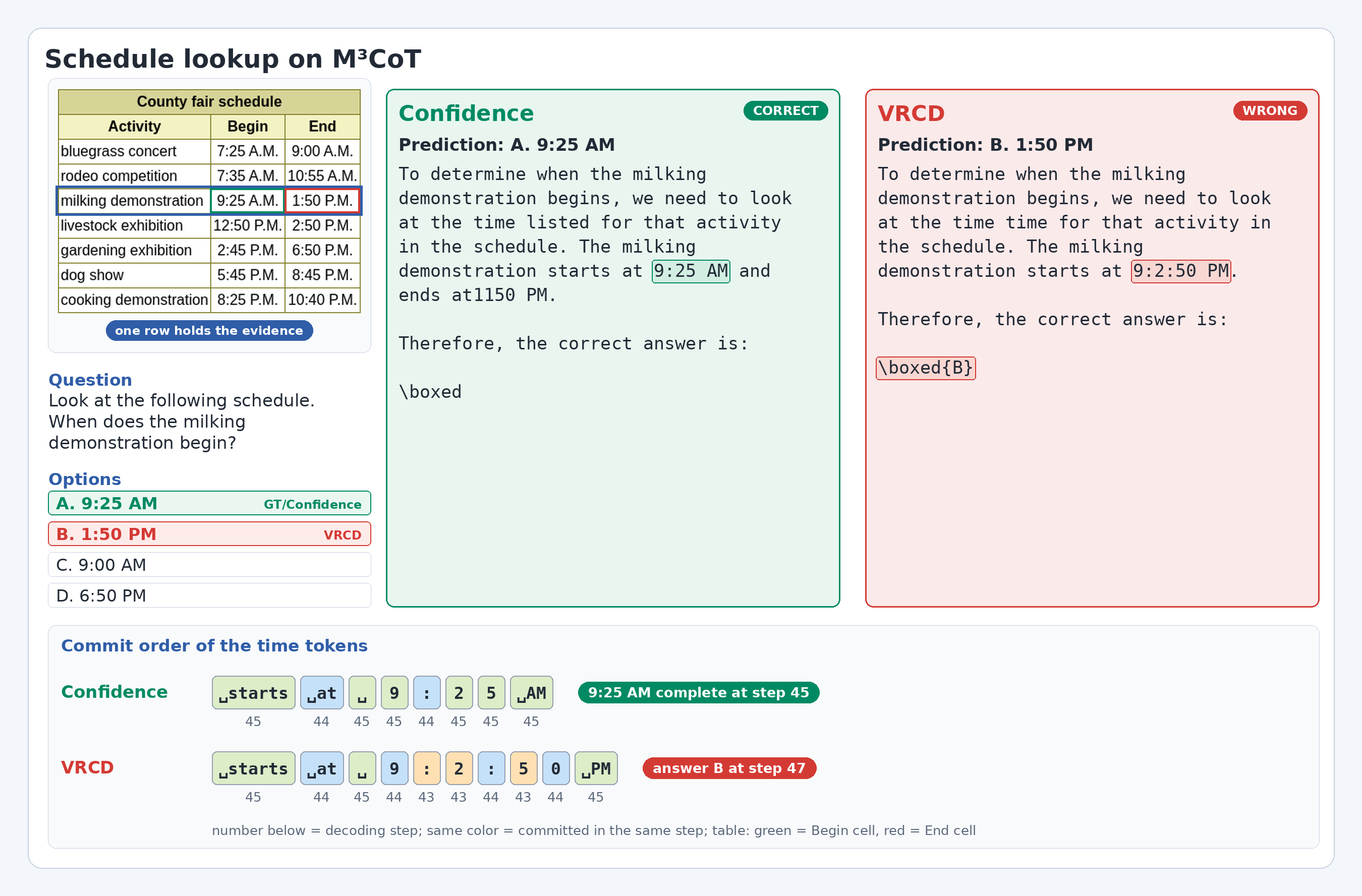}
    \caption{Schedule question from M$^3$CoT ($L=384$, $\mathrm{FR}=0.125$). The boxes mark the evidence row and its Begin and End cells. Bottom: the decoding step at which each token of the stated time is committed. Confidence-based unmasking commits 9, 2, 5, and AM at step 45; VRCD spreads the time over steps 43--45 and writes 9:2:50 PM.}
    \label{fig:case_failure}
\end{figure}

We present four cases: two MMBench examples that show where the outputs of the two methods begin to diverge (Fig.~\ref{fig:mmbench_case_study}), a case where lower overlap helps (Fig.~\ref{fig:case_positive}), and a failure case (Fig.~\ref{fig:case_failure}). In all case figures, green marks the correct option and the output that selects it, and red marks the incorrect option and the output that selects it. For the two M$^3$CoT cases, we also compare, at each step, the VRI of the tokens that VRCD commits with that of the top-$K_t$ positions by confidence in the same decoding state.

\paragraph{A case where lower overlap helps.}
\label{app:case_positive}
Fig.~\ref{fig:case_positive} shows an M$^3$CoT question that compares two gas samples. The particles in both samples have the same average speed, and those in Sample~A are heavier, so Sample~A has the higher temperature (option A). The example comes from the M$^3$CoT runs used for Tab.~\ref{tab:vri_inline} ($L=384$, $\mathrm{FR}=0.5$, 192 decoding steps). Confidence-based unmasking commits the answer B at step 123, before its text states either mass. Its later text concludes that Sample~A has the higher temperature, yet the committed B remains the final answer. VRCD first commits the digits of both masses (steps 124--129) and the statement that the two speeds are equal (steps 134--139), and it commits A at step 190. Across the decoding steps, the tokens that VRCD commits have a mean VRI of 0.590, against 0.698 for the top-$K_t$ positions by confidence, and the VRCD tokens have the lower VRI in every step that commits a mass digit (for example, 0.384 against 0.584 at step 125).

\paragraph{A failure case.}
\label{app:case_failure}
Fig.~\ref{fig:case_failure} shows an M$^3$CoT question that asks when the milking demonstration begins in a schedule table. The evidence lies in one row, and the answer is its Begin cell, 9:25 A.M. (option A). The setting is $L=384$ and $\mathrm{FR}=0.125$, as in Tab.~\ref{tab:main_len384}. Confidence-based unmasking commits ``9'', ``2'', ``5'', and ``AM'' together at step 45 and states 9:25 AM, which corresponds to A. VRCD commits the parts of the time over steps 43--45. In each of these steps, the tokens that VRCD commits have lower VRI than the top-$K_t$ positions by confidence (0.786 against 0.819, 0.698 against 0.733, and 0.783 against 0.790). Its text then reads ``9:2:50 PM'', which combines the Begin cell with the End cell 1:50 P.M., and VRCD selects B (1:50 PM). The tokens of one time value are grounded in the same cell and state different parts of one reading, so their overlap reflects necessary shared grounding (Appendix~\ref{app:overlap_boundary}). Lowering this overlap splits the reading across steps, and the image offers no complementary evidence to take its place.

\clearpage

\end{document}